%% file: ms.tex
\documentclass[10pt,technote]{IEEEtran}

\ifCLASSOPTIONcompsoc
  \usepackage[nocompress]{cite}
\else
  \usepackage{cite}
\fi

\usepackage{hyperref}

\ifCLASSINFOpdf
  \usepackage[pdftex]{graphicx}
\else
  \usepackage[dvips]{graphicx}
\fi

\usepackage{amsmath}

\begin{document}
\title{Measuring Human Perception to Improve Handwritten Document Transcription}

\author{Samuel~Grieggs,
        Bingyu~Shen,~\IEEEmembership{Student Member,~IEEE,}
        Greta~Rauch,
        Pei~Li,
        Jiaqi~Ma,
        David~Chiang,
        Brian~Price,
        and~Walter~J.~Scheirer,~\IEEEmembership{Senior~Member,~IEEE}%
\IEEEcompsocitemizethanks{\IEEEcompsocthanksitem S. Grieggs, B. Shen, G. Rauch, D. Chiang, and W. Scheirer are with the Department of Computer Science and Engineering, University of Notre Dame, Notre Dame, IN, 46556.\protect\\
Corresponding Author’s E-mail: walter.scheirer@nd.edu.
\IEEEcompsocthanksitem P. Li is with Zoom Video Communications.
\IEEEcompsocthanksitem J. Ma is with the Department of Classics, Yale University.%
\IEEEcompsocthanksitem B. Price is with Adobe Research.}%
}

\markboth{IEEE Transactions on Pattern Analysis and Machine Intelligence,~Vol.~X, No.~X, June~2021}%
{Shell \MakeLowercase{\textit{et al.}}: Bare Demo of IEEEtran.cls for Computer Society Journals}

\IEEEtitleabstractindextext{%
\begin{abstract}
In this paper, we consider how to incorporate psychophysical measurements of human visual perception into the loss function of a deep neural network being trained for a recognition task, under the assumption that such information can reduce errors. As a case study to assess the viability of this approach, we look at the problem of handwritten document transcription. While good progress has been made towards automatically transcribing modern handwriting, significant challenges remain in transcribing historical documents. Here we describe a general enhancement strategy, underpinned by the new loss formulation, which can be applied to the training regime of any deep learning-based document transcription system. Through experimentation, reliable performance improvement is demonstrated for the standard IAM and RIMES datasets for three different network architectures. Further, we go on to show feasibility for our approach on a new dataset of digitized Latin manuscripts, originally produced by scribes in the Cloister of St. Gall in the the 9th century. 
\end{abstract}

}

\maketitle

\IEEEdisplaynontitleabstractindextext

\IEEEpeerreviewmaketitle

\vspace{10mm}

\IEEEraisesectionheading{\section{Introduction}\label{sec:introduction}}

 In archives scattered around the globe, old manuscripts can be found piled up to the ceiling and spread out as far as the eye can see. The amount of writing produced on physical media since antiquity is staggering, and very little of it has been digitized and transcribed into plain text for researchers to study using modern data mining tools~\cite{terras:2012}. Work in the digital humanities has sought to address this problem by deploying everything from off-the-shelf optical character recognition (OCR) tools~\cite{boschetti2009improving} to state-of-the-art convolutional neural network-based transcription pipelines~\cite{firmani2018towards}. However, such work has been underpinned by the long-standing, yet incorrect, belief that computer vision has solved handwritten document transcription~\cite{wigington2018start,smith2018}. The open nature of this problem, coupled with a difficult data domain that has largely remained the realm of  specialist scholars, makes it a fascinating case study for testing the capabilities of artificial intelligence.
 
 For historical documents, work is ongoing in computer vision~\cite{wolf2011computerized,latin_comp,wigington2018start} and natural language processing (NLP)~\cite{berg2013unsupervised,smith2018} to produce a method that operates with little human intervention. However, to solve this problem, it helps to take a closer look at the behavior of reading. In this paper we seek out a fresh perspective by addressing three understudied aspects of the problem related to the way humans perceive handwritten documents: (1) the need in some cases to collect annotations for supervised learning from experts, (2) a loss formulation that makes use of behavioral measurements from people, and (3) a recognition pipeline that can successfully operate in a purely visual mode when it is not possible to train a language model on the statistics of language for post-correction.
 
\begin{figure}[t]
 \centering
 \includegraphics[width=.475\textwidth]{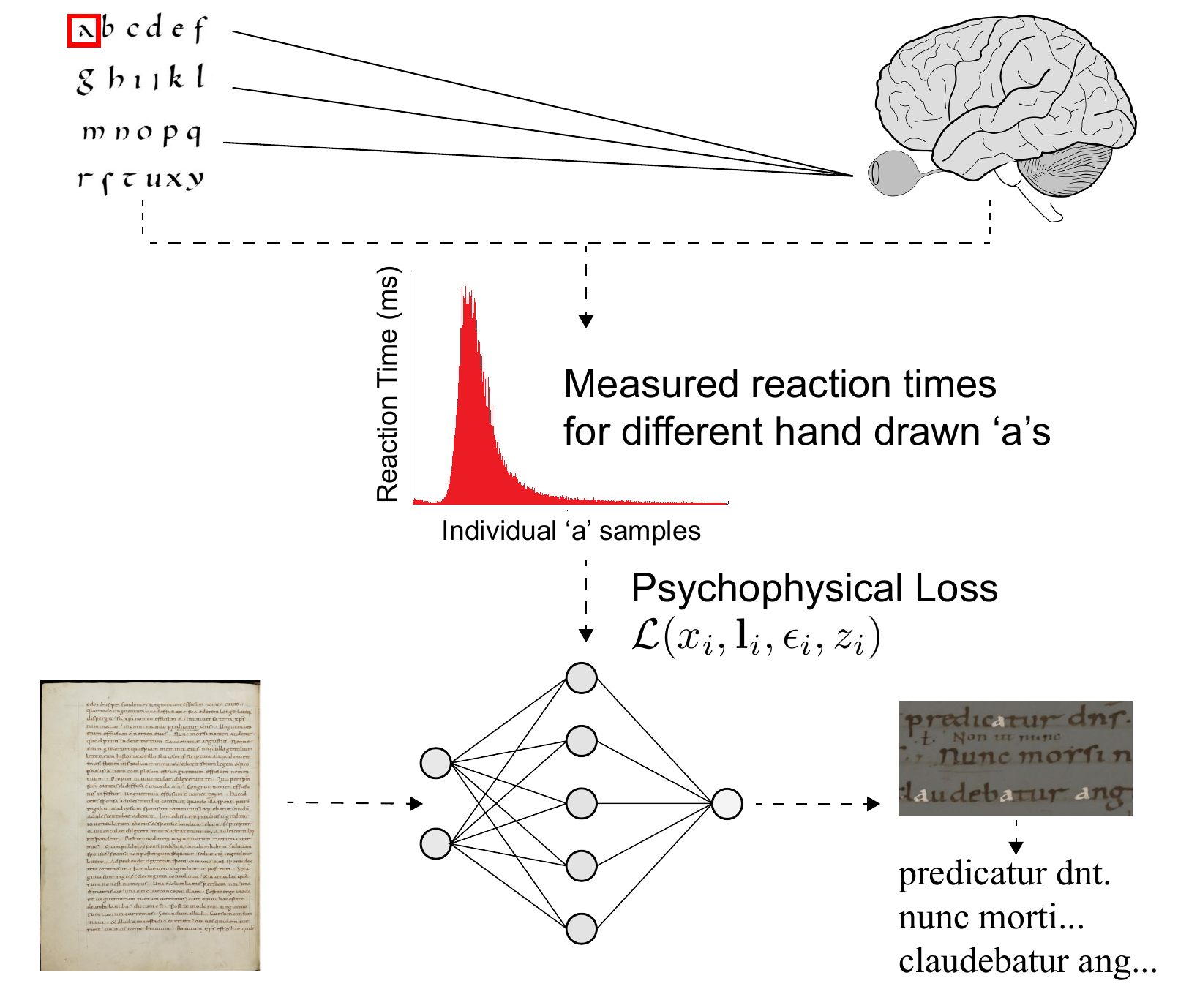}
 \caption{
 In this work we introduce a novel loss formulation for deep learning that incorporates measurements of human vision, which can be applied to different processing pipelines for handwritten document transcription.}
 \vspace{-3mm}
 \label{fig:teaser}
\end{figure}

The conventional wisdom in machine learning says that to improve performance on a task, more data should be collected. However, when it comes to document analysis tasks that require some form of expertise to produce supervised labels, it is not possible to turn to workers on Amazon's Mechanical Turk service to annotate a dataset. For example, if the task is to transcribe a parchment manuscript written in Latin, one will observe a large difference in behavior between the novice and experienced reader, with the novice producing an unreliable transcription. Researchers within the digital humanities have been struggling with the problem of a lower annotation yield for years~\cite{crane2012student}. In this work, we explore a new data annotation regime that allows us to collect more information from fewer annotators. The key is in the data we collect: images with more information-rich annotations.

Turning to the study of perception, annotations that are not basic labels, but instead measurements of the internal processes of the human visual system, have been shown to be effective for improving supervised machine learning~\cite{Scheirer_2014_TPAMI}. The strategy relies on the use of visual psychophysics, a set of methods and procedures from psychology that are used to study the relationships between physical stimuli and mental phenomena. For instance, the amount of time it takes a subject to recognize a handwritten character is correlated with the quality of the handwriting under scrutiny. Including this information into the training regime of a machine learning algorithm gives the resulting model some notion of what is difficult, and what is easy. This is critical for problems that contain images of varying degrees of difficulty. Previous work  has looked at conditioning classifiers on psychophysical data~\cite{Scheirer_2014_TPAMI}, but has not considered using such information for representation learning. Here we introduce a loss formulation for deep networks based on perceptual measurements (Fig.~\ref{fig:teaser}). 

The goal of current work in handwritten document transcription is to produce a useful tool that can generalize to new documents, given appropriate training data. Prior approaches have focused on individual aspects of the problem (\textit{e.g.}, image restoration~\cite{brown2007restoring}, character segmentation~\cite{firmani2018towards}, word recognition~\cite{poznanski2016cnn}, data augmentation~\cite{wigington2017data}, language modeling~\cite{wuthrich2009language}). However, NLP components that require dictionaries and other comprehensive corpora are problematic in some contexts. For instance, a separate neural language model is often trained to match the source language of the documents being considered. It takes the output of a vision-based transcription network and makes corrections based on a learned probability distribution of character- or word-level sequences. For historical documents, a  challenge exists in training language models under circumstances where incomplete knowledge of the lexicon is always present (\textit{e.g.}, a complete dictionary of Medieval Latin has never been compiled). In spite of this, we show that it is still possible to improve the performance of an imperfect vision system by making use of psychophysical measurements. 

\textbf{Contributions.} To summarize, the contributions of this paper are: (1) An open source software platform for collecting crowdsourced psychophysical and conventional annotations for historical documents, designed with the humanities scholar in mind. (2) A novel loss formulation incorporating measurements of human vision into the training process of three different deep neural networks. (3) A new collection of data for transcription evaluation including psychophysical annotations for partitions of the IAM~\cite{marti2002iam} and RIMES~\cite{augustin2006rimes} datasets, as well as a new dataset of open access Medieval Latin manuscript images and annotations. (4) Experiments showing that the proposed loss formulation reliably improves the performance of different network architectures on IAM, RIMES, and the new Latin manuscript set.

\section{Prior Work}

\textbf{Recent Advances in Document Transcription.} For general use, off-the-shelf OCR packages like Tesseract~\cite{smith2007overview}, OCRopus~\cite{breuel2008ocropus}, and ABBYY FineReader PDF~\cite{FineReader} remain popular. These packages are often the first choice of scholars working with historical documents because they are easy to use~\cite{springmann2014ocr}. However, because they have been designed for use with modern print documents, their performance is often very poor on documents outside of that domain. This can be seen in Table~\ref{tab:other_baselines}, which contains results for a transcription task that used the Latin manuscript dataset described in Sec.~\ref{sec:expsetup} to evaluate these packages. Tesseract in this case was trained with Latin data~\cite{LatinTess}, and FineReader configured to process Latin. Even with these capabilities, the performance is unacceptably low.   

The most important recent advances in document transcription are based on deep learning or probabilistic generative models. Poznanski and Wolf~\cite{poznanski2016cnn} introduced a feed-forward Convolutional Neural Network (CNN) that estimates the lexical attributes (\textit{i.e.}, n-gram features) of a word for transcription purposes. Doetsch et al.~\cite{doetsch2014fast} and Bluche et al.~\cite{bluche2014comparison} explored recurrent approaches for sequence processing by using Bidirectional Long Short-Term Memory (LSTM) in different input and architectural configurations. Bukhari et al.~\cite{bukhari2017anyocr} addressed the problem of labeled data scarcity in historical document datasets by using image segmentation and character clustering to automatically generate data for LSTM training. With respect to generative models, Berg-Kirkpatrick proposed jointly modeling the text of the document and the noisy process of rendering glyphs in an unsupervised manner~\cite{berg2013unsupervised}. The implementation is available as an open source software package called Ocular~\cite{ocular}, but is better suited to print documents (see Table~\ref{tab:other_baselines}). 

When it comes to the state-of-the-art on standard benchmark datasets for document transcription like IAM and RIMES, neural networks are combined with domain-specific heuristics for the target language and writing style. For instance, Multidimensional LSTM  (MDLSTM) networks have been paired with $n$-gram language models trained  over large-scale corpora containing a comprehensive lexicon~\cite{voigtlaender2016handwriting, puigcerver2017multidimensional}.
Some of the very best results on historical documents~\cite{german_comp,wigington2017data} have been achieved using the Convolutional Recurrent Neural Network (CRNN) architecture of Shi et al.~\cite{ShiBY17} and a Connectionist Temporal Classification (CTC) loss~\cite{graves2006connectionist} for training as a basis. The CRNN is able to support image-based sequence recognition in an end-to-end trainable fashion. Xiao et al.~\cite{xiao2019deep} proposed a number of improvements to CRNN including rectification that corrects the character distortions that are prevalent in the IAM dataset, as well as post-processing by a word beam search language model. We will show that training with our proposed loss function can improve CRNN performance without comprehensive knowledge of the target language or changes to the architecture.

\input{other_models.tex}

\textbf{Computer Vision for Manuscript Studies.} Images of Medieval manuscripts provide a unique challenge for computer vision algorithms, given the prevalence of poor image quality, inconsistent handwriting, and often incomplete knowledge about the language at hand. Work on this problem is highly interdisciplinary, where the end goal is to provide a useful tool to scholars in the humanities. Thus engineering has gone into creating end-to-end processing pipelines that are accessible to paleographers~\cite{wolf2011computerized,firmani2018towards,abbott2017time,kahle2017transkribus,Stutzmann2018}. Early work in computer vision for manuscript studies articulated the problem as one of sequence processing over lines from the images, where approaches like HMMs apply if trained on the target language~\cite{edwards2005making,fischer2011transcription}. Secondary problems in data quality were noted, and various approaches to fix them have been proposed~\cite{brown2004image,brown2007restoring,lu2009directed}.  

As in automated transcription for modern texts, neural network-based approaches are now routinely deployed for manuscript analysis. Fischer et al.~\cite{fischer2009automatic} demonstrated the effectiveness of a Bidirectional LSTM in transcribing a 13th century manuscript written in Middle High German. Firmani et al.~\cite{firmani2018towards} showed over a corpus of manuscripts from the Vatican Apostolic Archive that if character segmentation is accurate, a CNN can be used to classify characters with high accuracy. Related work was undertaken to classify Medieval handwriting for archival cataloguing~\cite{latin_comp}.  But error rates remain stubbornly high, thus leaving humans in the loop to make corrections.  

\textbf{Visual Psychophysics for Computer Vision.} 
Work in this direction has looked at the sensitivity of different visual systems to the local statistics of natural images~\cite{gerhard2013sensitive}, the behavioral consistency of deep networks to human vision~\cite{eberhardt2016deep,Taylor_2020_CVPR_Workshops}, and human vs. non-human primate vs. deep network object recognition performance~\cite{DBLP:journals/corr/GeirhosJSRBW17, rajalingham2018large}. A psychophysics-inspired evaluation framework has been proposed for object recognition in computer vision~\cite{RichardWebsterPsyPhy2018}. Most closely related to the work in this paper is that of Scheirer et al.~\cite{Scheirer_2014_TPAMI}, where a loss function that applies penalties based on psychophysical annotations attached to individual examples during classifier training is used to improve classification accuracy. 
In contrast to classification methods that gradually emphasize more difficult examples like boosting~\cite{schapire2003boosting}, a psychophysical loss function exploits precise measurements of human vision that are common in psychology, but not in computer vision. 

There are four key distinctions between our work and that of Scheirer et al.~\cite{Scheirer_2014_TPAMI}. (1) We extend their strategy, which was limited to Support Vector Machine (SVM) classifiers, to representation learning by introducing a loss formulation for deep networks that makes use of psychophysical measurements.  (2) The proposed loss function is more sophisticated than the loss function introduced by Scheirer et al., operating in two modes depending on what type of examples (easy or hard) should be prioritized during training. (3) This work looks at the collection of psychophysical measurements from experts (in this case readers of Latin), as opposed to non-experts on a crowdsourcing platform. (4) This work concentrates on the specific application of handwritten document transcription, while prior work has limited itself to face detection.

\section{A Loss Formulation That Makes use of Measurements from Human Readers}
\label{sec:method}
The approach for handwritten document transcription we consider assumes an end-to-end pipeline composed of a set of distinct modules for computer vision and natural language processing. We begin with a data collection process incorporating expert readers making use of a custom psychophysical data collection tool. This provides data that can be used by a standard handwritten text recognition model, but also allows us to collect additional data that measures the process of reading. These psychophysical measurements of human vision can be incorporated into a loss function that improves performance. Here we describe the new loss formulation, explaining how to go from psychophysics experiments to a training strategy for any CTC-based transcription pipeline\footnote{All code and data will be released after  publication.}.

\textbf{Psychophysical Data Collection.} Historical documents present unique challenges for handwritten document transcription algorithms. For instance, while there are thousands of high-quality images of Medieval manuscripts available on the Internet, very few of those documents have been annotated for transcription tasks. Even more problematic, such documents cannot be annotated by just anyone --- paleographic experience is often required. In a pilot study, we had a reader who was unfamiliar with the Latin language transcribe a manuscript page from the 9th century. This reader's result was scored against the transcription produced by a knowledgeable reader of Latin, and had a character error rate of $35\%$. Therefore, this task cannot simply be farmed out to Amazon's Mechanical Turk service. It requires significant subject matter knowledge.  Furthermore, even when an expert does take the time to transcribe manuscript images, they often make editorial changes to expand abbreviations, and correct spelling errors and other abnormalities in the text, making the resulting ground-truth transcription unsuitable for training a system that can produce an exact (\textit{i.e.}, diplomatic) transcription. 

To address these challenges we developed a web-based tool to leverage expert reading behavior that not only produces exact ground-truth data, but also collects psychophysical measurements of the process of reading. The tool is based on a version  of  the  Image  Citation  Tool  from  the  Homer Multitext  Project~\cite{blackwell2014}, which provides a familiar interface to non-technical annotators working on digital data curation for historical documents. An additional stage was added to collect psychophysical annotations. In terms of the workflow, an expert reader is shown a screen in which they are prompted to identify a single character that has been hand-segmented from the page. The reader is then timed as they select which character is represented in the image from a pre-defined set of characters that cover all of the possibilities for a dataset. Each character can be shown to multiple readers, with their reaction times averaged to account for environmental differences. 

This type of psychophysical data collection is not just constrained to expert reading problems. Psychophysical annotations for modern languages such as English or French can be collected in a manner similar to the above process, but with a bit more flexibility. Since most people who are fluent in a Western language can easily read print and cursive handwriting, there is no need to present them with segmented characters. For modern languages, the reader types out a candidate line as they parse it. While each line is being transcribed, the individual keystrokes were timed and recorded. The data are collected in small batches of 25 images, with 5 control images that the rest of the measurements are normalized against to control for environmental factors across readers, such as different typing speeds. 
Supp.~Mat.~Sec.~1 contains additional details and screenshots of the platform.

\textbf{Psychophysical Loss Function.} The tools described above can measure timing information at the character- or line-level. In this work, we choose to operate at the line-level with respect to input to the transcription system. Character-level measurements are averaged across a line to normalize for length. Thus each line image has an associated average reaction time $r$ (see Supp.~Table~1). This form of psychophysical measurement is inspired by the work of Scheirer et al.~\cite{Scheirer_2014_TPAMI},
which demonstrated that utilizing reaction time as a psychophysical measurement can improve the performance of SVM. Other psychophysical measurements could be used as well, including accuracy (if  ground-truth is available before the behavioral measurements are taken) and eye movements.  

In this work, we propose a loss formulation for training CNNs to perform handwritten document transcription. In order to facilitate this, we collect psychophysical measurements from a subset of lines from each dataset and use them to derive penalties that will be applied by the loss function during training. In practice, these measurements can help the network prioritize the importance of examples that are easier or harder for humans to transcribe, depending on the application.  

\begin{figure}[t]
 \centering
 \includegraphics[width=.475\textwidth]{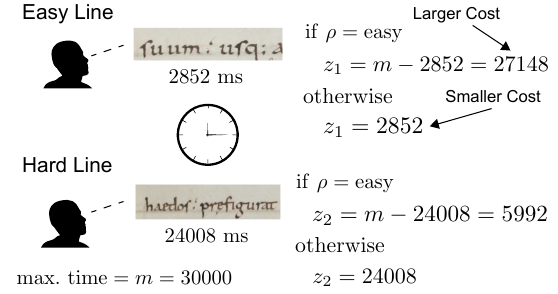}
 \caption{An example of how the psychophysical penalty $z_i$ is calculated when the priority ($\rho$) is on easy or hard examples. %
 }
 \vspace{-3mm}
 \label{fig:penalty}
\end{figure}

In order to make use of
the psychophysical data that was collected, we designed a new loss function that is a variation
of the commonly used CTC loss function originally introduced by Graves et al.~\cite{graves2006connectionist}. This is the standard loss function for transcription tasks. CTC loss is the negative logarithm of the probability of the ground-truth given a predicted text sequence. After calculating this loss for a specific training image, our method adds a psychophysical penalty that is scaled to the character error rate of the output. 

Assume we have a training data set ${\cal X} = \{x_i, \mathbf{l}_i, \epsilon_i, z_i \}$, where $x_i$ is the training image, $\mathbf{l}_i$ is the associated ground-truth label sequence, $\epsilon_i$ is the associated character error rate, and $z_i$ is an associated psychophysical penalty. The objective when using CTC loss is to minimize the negative log-likelihood of the conditional probability of the ground-truth:  
\begin{equation}
{\cal L}_\mathrm{CTC} = -\sum_{x_i, \mathbf{l}_i \in {\cal X}} \mathrm{log}~p(\mathbf{l}_i | \mathbf{y}_i)
\label{eq:ctc}
\end{equation}
where $\mathbf{y}_i$ is the sequence produced by the recurrent and convolutional layers of a network when processing $x_i$. 

To reformulate the above loss to incorporate the psychophysical measurements, we must first calculate a psychophysical penalty for each image that matches our assumptions about visual perception and reading (Fig.~\ref{fig:penalty}). Assume that we have a  set $R = \{r_1, \ldots, r_n\}$ of sorted average reaction time measurements from an experiment. The maximum value of that set is $m = \mathrm{max}(R)$. Also assume that images with a short measured reaction time (in ms.) are easy for a human observer to recognize characters in, and images with a long measured reaction time are more difficult.  To calculate a psychophysical penalty $z_i$ that prioritizes easy examples during training, an average reaction time $r_i$ is subtracted from $m$. To prioritize hard examples (as in hard example mining~\cite{Shrivastava_2016_CVPR,Lin_2017_ICCV}), $z_i$ is simply assigned to $r_i$.  
\begin{equation}
    z_i =
    \begin{cases}
      m - r_i, & \text{if~easy~examples~prioritized} \\
      r_i, & \text{otherwise}
    \end{cases}
\end{equation}

A loss function conditioned on the psychophysical measurements should have the effect of pushing the training in a direction that will perform better on the prioritized examples, with priority specified by a parameter $\rho$. It should also preserve the penalties for incorrect predicted text sequences. Thus we combine the CTC loss (as defined in Eq.~\ref{eq:ctc}) with an additional loss term derived from the psychophysical measurements and the character error rate associated with an image: 
\begin{equation}
{\cal L}_\mathrm{psych} = -\sum_{x_i, \mathbf{l}_i, \epsilon_i, z_i \in {\cal X}} \left( \mathrm{log}~p(\mathbf{l}_i | \mathbf{y}_i) + (-\epsilon_i \times z_i)\right)
\end{equation}

\section{Experimental Setup}
\label{sec:expsetup}
\textbf{Datasets.} As a baseline for assessing the effect of training with the proposed loss function, we consider two standard databases: IAM~\cite{marti2002iam} for English and RIMES~\cite{augustin2006rimes} for French. The IAM database contains more than $10,000$ images of handwritten lines of text created by $657$ writers. The texts those writers transcribed are from the Lancaster-Oslo/Bergen Corpus of British English~\cite{corpuslancaster}. The dataset has been split according what is known as the Aachen split, specifically the Aachen split as defined by Puigcerver et al. in~\cite{laia2016}. It has a training set ($6,161$ lines), a validation set ($966$ lines), and a test set ($2,915$ lines). The RIMES database consists of images of lines of text from letters composed by volunteers. While the dataset does not have an official training and validation split, we used the same split described by Puigcerver et al.~\cite{laia2016}, which is standard practice in the literature. This split has a training set ($1,0171$ line), validation set ($1,161$ images), and testing set ($778$ images).
For the challenging historical documents that we are primarily concerned with, we  created a new set of open access Latin language documents. Our Latin dataset comes from e-codices, which is the Virtual Manuscript Library of Switzerland~\cite{ecodices}. The specific codex selected is from the Cloister of St. Gall~\cite{stgall159}. We collected $1,861$ images of handwritten lines that were split into a training set ($1,479$ images), a validation set ($196$ images), and a test set ($186$ images) in a page disjoint fashion. It is not known how many writers this set consists of, but in practice, all scribes wrote in a deliberately similar script.

We collected psychophysical annotations for $2,132$ text lines from IAM, which is approximately $35\%$ of the training set. For RIMES, we collected psychophysical annotations for $1,257$ lines, which is approximately $12\%$ of the training set. These numbers reflect the amount of data that we were able to collect through psychophysics experiments during the time period this research was conducted. The psychophysical annotations for IAM were collected from graduate students fluent in English. The RIMES annotations were collected from graduate students with a working knowledge of French.
With respect to the Latin dataset, the texts were annotated by graduate students who were fluent in Latin and trained in paleography. %
We asked our annotators to create exact transcriptions from the lines they were provided. This dataset has psychophysical annotations across the entire training set of $1,479$ images. 

\textbf{Handwritten Text Recognition Networks.} In order to demonstrate improvement across different neural network architectures, we ran our experiments with three different handwritten text recognition networks. The first is the standard CRNN architecture. The implementation we used was identical to the one used by Puigcerver in~\cite{puigcerver2017multidimensional}. It consists of five convolutional layers and three bidirectional LSTM layers. The second architecture was a variation on the CRNN, where the convolutional layers were replaced with a U-Net~\cite{ronneberger2015u}  segmentation module. Because of the added complexity of adding the U-Net, the number of LSTM layers was reduced to two. We refer to this architecture as URNN.  The third architecture was also a variant of CRNN, where the recurrent layers were replaced with a transformer encoder~\cite{vaswani2017attention} with eight heads. This model is referred to as CTransformer. Additional details for these architectures can be found in Supp.~Mat.~Sec.~2.

For the CRNN and CTransformer networks, we generated five random initializations and trained each network with both the standard CTC loss function and our proposed psychophysical loss. The URNN model takes an order of magnitude longer to train, so we followed the same procedure, but with only three random initalizations. Each experiment uses the same training configuration, stopping after $80$ epochs without improvement. For the CRNN and CTransformer, we cut the learning rate in half every $15$ epochs without improvement, but we found that this made the URNN fail to converge, so we increased this interval to $30$ epochs for that network. For each architecture we used the optimizer found to be most effective, which was RMSProp for CRNN, Adadelta for URNN, and Adam for CTransformer. A difference from other setups found in the literature for the experiments performed on IAM and RIMES is that we reduced the image height to $64$ from the typically used $128$ to increase processing speed. This had a slight negative impact on overall performance, but did not affect the assessment of the loss function effect. 

\section{Experimental Results}
\input{avg_res_val_tab.tex}
\textbf{Measurable Effect of the Psychophysical Loss.} The primary set of experiments assessed the change in network performance after the move to the new loss function. An overview of the results with priority on easy examples can be seen in Tables~\ref{tab:average_results_val} and~\ref{tab:average_results_test}. These tables show the average results for each set of experiments by model and dataset.  A complete set of the $88$ individual experiments that were run is included in Supp.~Mat.~Sec.~3. We also include tables in the Supp. Mat. that show the best result from each experiment. With respect to the overall measurable effect, we find that the psychophysical  loss function with priority on easy examples gives a consistent (across datasets and architectures) and repeatable (across trials within a dataset for an architecture) improvement to handwritten text recognition performance. 
\input{avg_res_test_tab.tex}

Across the $44$ pairs of experiments we performed on identical initializations with and without the psyschophysical loss function, we found that the models trained using the new loss preformed better (in terms of Test Set CER) set in $40$ of them. Every error rate is improved on average with the psychophysical loss, except for the WER of the CTransformer on the Latin validation set. It is also worth noting that the best overall result on the test sets from each set of experiments was always from a model trained using the psychophysical loss function (see Supp. Tables 2, 3 and 4). 

Experiments were also performed putting the priority on hard examples for the psychophysical loss function (Supp. Tables 7 \& 8). The results for this mode of the proposed loss function are mixed compared to putting the priority on easy examples. In some cases lower error rates are achieved (\textit{e.g.}, CRNN on all datasets), but in other cases the error rates are higher than the baseline (\textit{e.g.}, CTransformer on all datasets). One possible explanation for the consistent improvement observed for easy priority may be the normalizing effect induced by subtracting every measurement from the maximum average reaction time. Another possible explanation is that the hard examples are not as reliably correctable as the easy examples during training. The loss function priority should be treated as a hyperparameter, with a decision on the priority made after evaluating both modes using a validation set.

There are two key differences in these experiments compared to other recent work in document transcription. The results shown in Tables~\ref{tab:average_results_val} and~\ref{tab:average_results_test} do not make use of a language model as part of the transcription pipeline. The primary reason for excluding a language model is that due to our specific interest in historical documents with large amounts of novel language content, it is not possible to train an effective one. Further, these experiments do not make use of any pre-training on synthetic data, since it is not possible to gather psychophysical annotations for synthetic handwritten text images generated on the fly. Nevertheless, the raw improvement counts are substantial, and represent a marked improvement for the reader of the transcriptions. Qualitative examples can be found in Supp.~Mat.~Sec.~5. 

Given the frequent need in the digital humanities to transcribe very large archives of documents, we also made some decisions that optimize for speed rather than overall performance. This helped in processing the large number of folds run for these experiments. As mentioned earlier, we used an image height of $64$ pixels rather than the more standard $128$. This had a notable impact on reducing computation time. Using a smaller network size to process smaller input images cuts the computation time from approximately $600$ seconds per epoch to about $250$. At inference time, the $64$ pixel input height allows for the processing of $31$ images per second, while the $128$ pixel input height yields $16$ images per second. These numbers were calculated using one NVIDIA Titan Xp GPU.  %

\textbf{Control Experiments.} 
We performed a series of control experiments to further verify that psychophysical measurements carry a useful signal for machine learning (with and without the proposed loss function), and to investigate if the proposed loss function is useful for penalties derived from data other than psychophysical measurements. All of the following experiments were run from the same five random initializations of the CRNN architecture from the set of experiments described above for the IAM dataset. Full details for the setup of each of these experiments can be found in Supp.~Mat.~Sec.~6. All results are in Table~\ref{tab:control}, with the baseline results in the row labeled `B', and the results using the psychophysical loss with psychophysical measurements labeled `E' for priority on easy examples and `H' for priority on hard examples.  

The first control experiment examined the use of algorithmic hard example mining instead of psychophysical measurements using the proposed loss function. Hard examples have proved to be useful in other vision applications (\textit{e.g.},~\cite{Shrivastava_2016_CVPR,Lin_2017_ICCV}), and their compatibility with the proposed loss formulation makes them an interesting test case. To do this, we took the baseline CRNN results from each fold, averaged the character error rates for each image, and used those averages as the associated penalties for the images. The results (the row labeled `M') show that by using hard examples, improvement over the baseline CRNN (`B') is possible. However, next to the new loss function with psychophysical measurements (`E' \& `H'), the results are comparable (\textit{i.e.}, within the error of `E' and close to `H') for CER, and a bit worse for WER. Thus the new loss works for data other than psychophysical measurements. 

The second control experiment assessed the utility of psychophysical measurements in a curriculum learning regime~\cite{bengio2009curriculum}, which also considers example difficulty. A curriculum was defined by taking all of the available psychophysical measurements for the IAM dataset, sorting them by reaction time per character, and then dividing them into five groups based on this ordering of difficulty. Groups were incrementally introduced in training after performance on the validation set did not improve for 20 epochs. The results (the row labeled `C') show improvement over the baseline CRNN results (`B'). This demonstrates that the psychophysical measurements can help machine learning training without the need of a specialized loss function. The results are also comparable to the psychophysical loss function (`E' \& `H'), which is promising for future work combining psychophysics with machine learning.

The third control experiment assessed another measure of example difficulty. We replaced the psychophysical measurements for IAM with the text length of each example (shorter assumed to be easier). We then trained using the psychophysical loss as we did for the experiments reported for the psychophysical data. The results (the row labeled `L') show some improvement over the baseline CRNN (`B') --- but not significant improvement. The results from this measure were not as good as those using psychophysical measurements (`E' \& `H'), or the hard example mining approach (`M').

\input{table_control}

\textbf{Larger Input, Language Model and Synthetic Pre-Training.} 
While the previously described networks trade-off speed for accuracy by using a small image height, their performance gets closer to the current state-of-the art results on modern language datasets when using a larger image height (128 pixels) and adding language model post-correction and synthetic pre-training. Language models are an important strategy for achieving good performance on modern language datasets like IAM, but do not work well for languages where we lack a comprehensive lexicon --- like Medieval Latin. This is not our primary use case, but is included for comparison.

The language model we used is a word beam search~\cite{scheidl2018wordbeamsearch}. Synthetic data for the transcription network was generated using the same methodology as Xiao et al.~\cite{xiao2019deep}. Training details for the language model and transcription network can be found in Supp.~Mat.~Sec.~7. As can be seen in Table~\ref{tab:comparison}, our approach (hard priority `H') yields the third best overall results on the IAM validation and test sets. This is achieved with a far simpler architecture (the original CRNN) than the approach proposed by Xiao et al. The proposed loss reduced error as expected (with easy `E' and hard `H' priority). 

\input{table_comparison}

In Table~\ref{tab:comparison}, comparison results are reported as they appear in published papers. This means that things like the language models used for post-correction and pre-possessing steps differ between approaches. Also, with the exception of the experiments reported by Puigcerver et al.~\cite{puigcerver2017multidimensional}, error bars are not reported, making it difficult to tell what the effect of changing the seed for initialization is (a major motivation for the use of multiple folds in our experiments above). For instance, our CRNN model without using the psychophysical loss is identical to the ``baseline'' and ``Recogm'' models used in the ablation studies by Puigcerver et al. and Xiao et al.~\cite{xiao2019deep} --- but all three implementations differ somewhat in performance.

\section{Discussion}

The motivation for this work has been the general need for a useful handwritten document transcription tool for the user who is not familiar with computer vision, but may be an expert in another field looking at challenging or unusual images. What did we do differently than previous approaches to get closer to fulfilling this need? In essence, we studied the reading behavior of experts using the methods and procedures of visual psychophysics from the field of psychology. By measuring the time it took experts to recognize particular characters, we gained a better sense of which ones are easy to recognize and which are more difficult. Considering historical documents, small, yet consequential, mistakes can completely alter the meaning of important lines, possibly leading to misinterpretations of literature and history. Avoiding these mistakes is essential for making transcription tools more useful. 

To operationalize the above idea, we avoided  creating a transcription pipeline for a specific language or style of writing, and instead opted to formulate a general loss strategy for CNNs. We demonstrated that collected psychophysical measurements could be incorporated into the training regime of a character recognition network to improve the accuracy of the transcriptions.
With respect to the evaluation of neural network-based handwritten document transcription pipelines, we suggested a strategy that makes use of multiple folds with different random seeds as a form of sensitivity analysis, showing before and after results for changes to the loss function across different languages and many different writers. We also recommended decoupling the vision and language components of transcription pipelines, because for many historical documents the lack of complete dictionaries for the target language leads to less effective language models.  

Even with these advances, there is still a ways to go in the development of an off-the-shelf tool like Tesseract for this problem. Subtle shifts in orthographic conventions within a language create ambiguity for current segmentation models. An alternative approach is to make pixel-level annotations available at training time, similar to segmentation in medical imaging applications.  This might provide more information from which to disambiguate characters.  With respect to language model improvements, there is a need to handle the abbreviations that are common for certain historical periods. In Latin manuscripts, paleographers expand these to the original word when producing transcriptions. For example, in Fig.~\ref{fig:teaser}, the abbreviation \textit{dnt.} should be expanded to the word \textit{dicunt}. A language model that can do this automatically is needed. 
Quite to the contrary of common belief, there is still significant and challenging work to be done on handwritten documents.

\ifCLASSOPTIONcompsoc
  \section*{Acknowledgments}
\else
  \section*{Acknowledgment}
\fi

This work was supported in part by the following organizations within the University of Notre Dame: Notre Dame Research, the College of Arts and Letters, the Medieval Institute, the Office of Mission Engagement, and the Office of Digital Learning. Additional support was provided by Adobe Systems and the NVIDIA Corporation. We thank Cana Short, Mihow McKenny,  Melody Wauke, Curtis Wigington and Christopher Tensmeyer for their help preparing data and invaluable advice. 

\ifCLASSOPTIONcaptionsoff
  \newpage
\fi

{\small
\bibliographystyle{IEEEtran}
\bibliography{egbib}
}

\end{document}


\title{Supplemental Material: Measuring Human Perception to Improve Handwritten Document Transcription}
\date{}

\maketitle

\section{Data Annotation Platform}
\begin{figure}[h!]
 \centering
 \includegraphics[width= .49\textwidth]{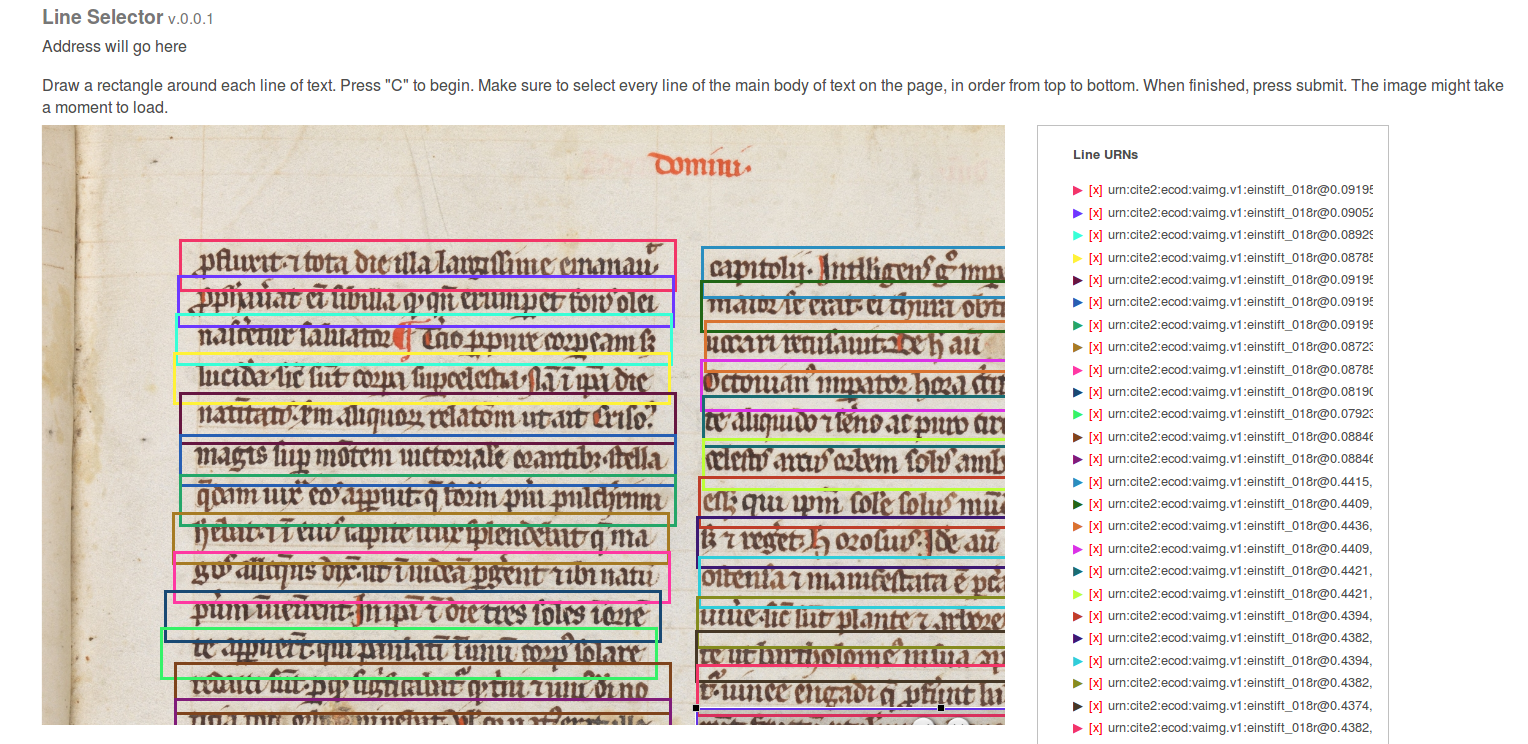}
 \includegraphics[width= .49\textwidth]{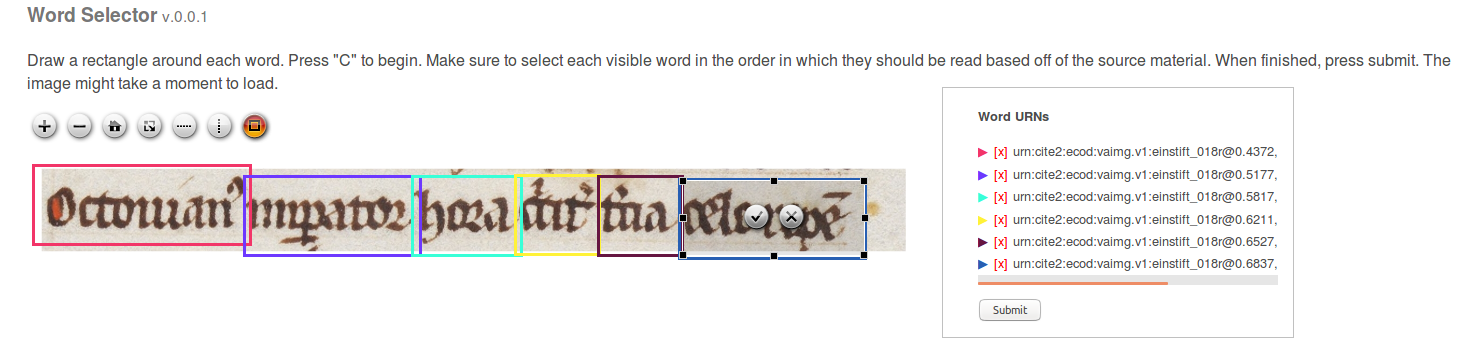}
 \caption{A screenshot of the line and word selection feature of our annotation platform being used with a document from the e-codices archive (Einsiedeln, Stiftsbibliothek, Codex 629(258), f. 4r – [Jacobus de Voragine] \textit{Legenda aurea sive lombardica}, \url{http://www.e-codices.unifr.ch/en/list/one/sbe/0629})}
 \label{fig:sup_lineswords}
\end{figure}

 Here we provide some additional details about, and some screenshots of, our data annotation platform. The platform is very user-friendly, and can be accessed via a web browser. We built the tool on top of the Homer Multitext Project's Image Citation tool~\cite{blackwell2014}, which is used to generate CITE2 URNs\footnote{\url{http://cite-architecture.org/cite2/}} that can be used to reference specific portions of scanned document images. The transcriber segments the document into lines and words (Fig.~\ref{fig:sup_lineswords}), and the tool generates the appropriate CITE2 URNs that reference the associated regions of the images. This process facilitates scholarly archiving for projects involving collections of digital artifacts, as well as provides us with the training data necessary for a deep learning-based handwritten document transcription pipeline. 
 
\begin{figure}[h!]
 \centering
 \includegraphics[width= .7\textwidth]{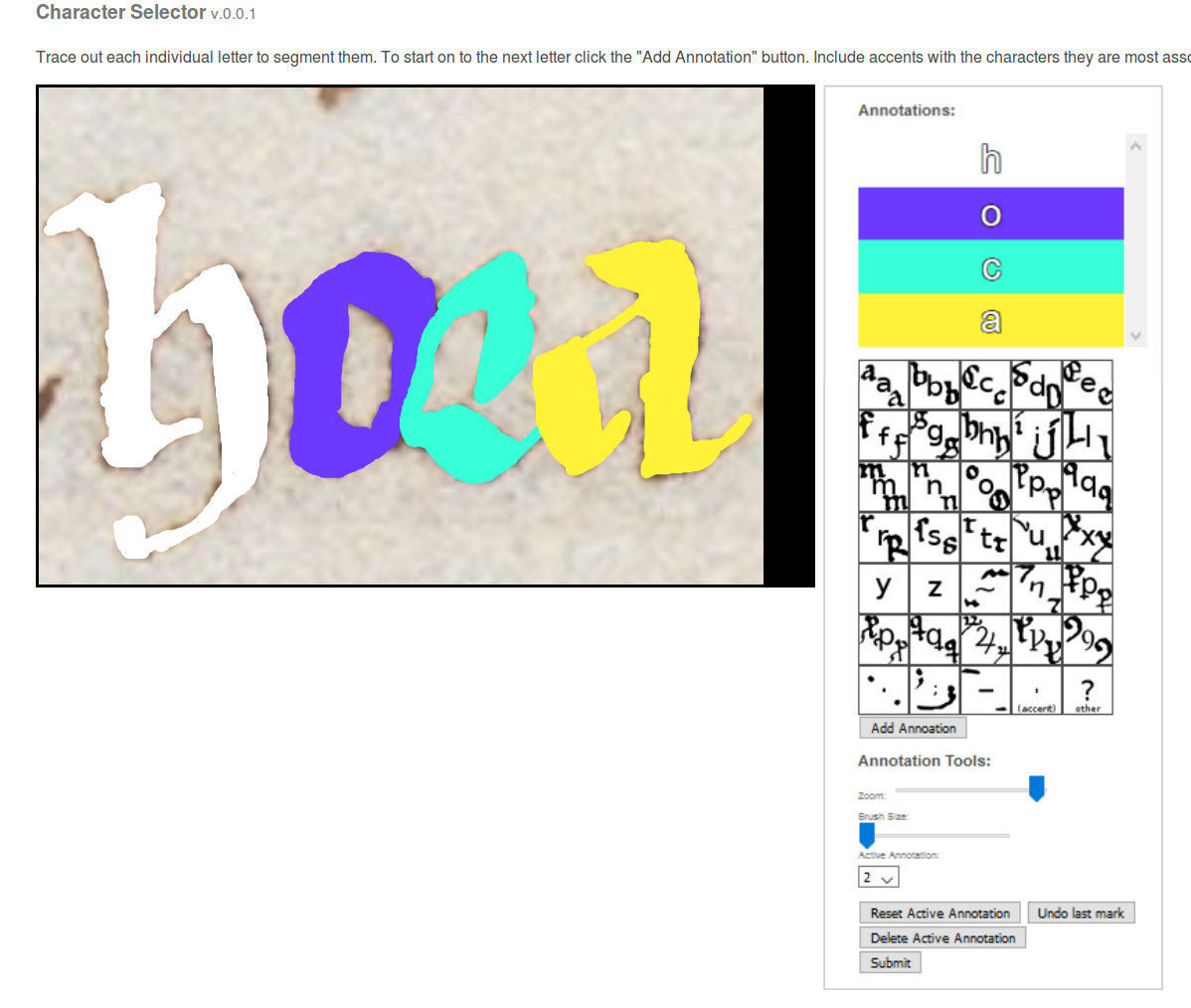}
 \caption{A screenshot of a word being traced out using our annotation platform.}
 \label{fig:word}
\end{figure}

After segmenting the document into words, the platform prompts the user to segment and annotate each word letter by letter.  Instead of using a bounding box, we have the user trace over each character in the word using a pen tool. This gives us a pixel-by-pixel segmentation mask of the document image. At this stage the user (leaning on any expert knowledge they possess) will also select which letter best represents each character from an array of buttons representing a predefined list of letter classes, as shown in Fig.~\ref{fig:word}.

\begin{figure}[h!]
 \centering
 \includegraphics[width= .75\textwidth]{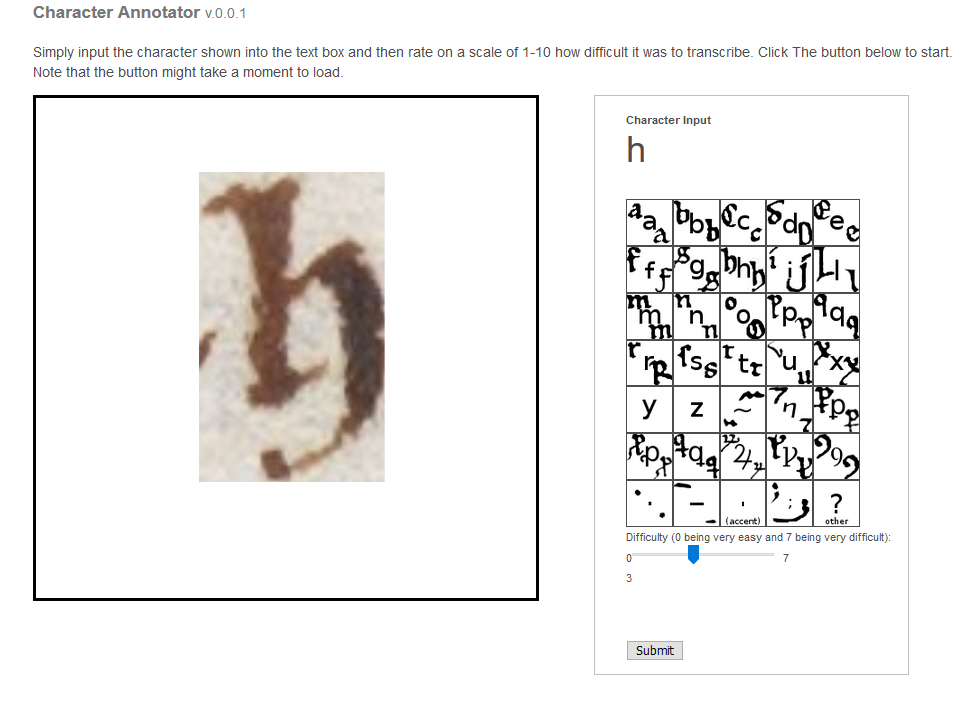}
 \caption{The tool used to measure the
performance of trained paleographers on Medieval manuscripts. It keeps track of reaction time as an annotator performs the task.}
 \label{fig:letter}
\end{figure}

\begin{figure}[t]
\begin{center}
 \includegraphics[width=.9\textwidth]{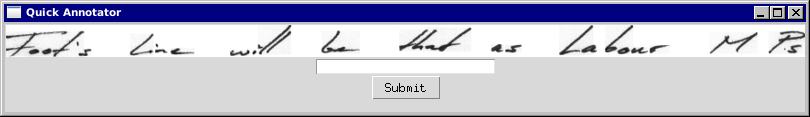}
 \caption{The tool for measuring reading performance for modern handwritten languages (\textit{e.g.}, the text in IAM and RIMES). Since  most  people  who  are fluent in a Western language can read print and cursive handwriting, there is no need to present them with  segmented characters. For modern languages, the reader types out a candidate line as they parse it. While each line is being transcribed, the individual keystrokes were timed and recorded.}
 \label{fig:annotations}
\end{center}
\end{figure}

The final stage collects the psychophysical measurements of reading. The platform brings up individual characters, as shown in Fig.~\ref{fig:letter}, and asks the transcriber to pick a label for a character without word-level context. The software times the user and also asks them to objectively state how difficult a particular character was to read.

The annotation platform also possesses a simpler tool to more quickly collect data from existing datasets, as shown in Fig.~\ref{fig:annotations}. It displays an image and asks the user to type out any text contained within it.  This feature times how long it takes the reader to type each line, and keeps track of the keystrokes logged in the process of typing out the text. Example lines from each dataset that the human annotators found easy and difficult can be seen in Table~\ref{tab:dataset_ex}.

\input{table1_examples_of_lines}

\newpage

\section{CRNN, URNN and CTransformer Diagrams} 

Included here are diagrams that describe the CRNN, URNN, and CTransformer architectures that were used for the experiments in this paper. CRNN was originally proposed by Shi et al.~\cite{ShiBY17}, and modified versions were also used by Puigcerver et al.~\cite{puigcerver2017multidimensional} and Xiao et al.~\cite{xiao2019deep}. Ours is structurally the same as the architecture used by Puigcerver et al. URNN and CTransformer are both modified versions of the CRNN architecture. 
In URNN, the convolutional layers at the beginning are replaced with a U-Net, which is a segmentation model originally proposed for biomedical image segmentation by Ronneberger et al.~\cite{ronneberger2015u}. The network is structured such that the U-Net attempts to predict which character class each image segment belongs to, and then passes that information into the bidirectional LSTM layers to refine the prediction with the surrounding context.
Similarly, CTransformer is a CRNN, but the LSTM layers are replaced with Transformer encoders, as specified by Vaswani et al.~\cite{vaswani2017attention}. The transformer layers serve the same purpose as the LSTM layers do in the original model.

\begin{figure*}[h]
 \centering
 \includegraphics[width=\textwidth]{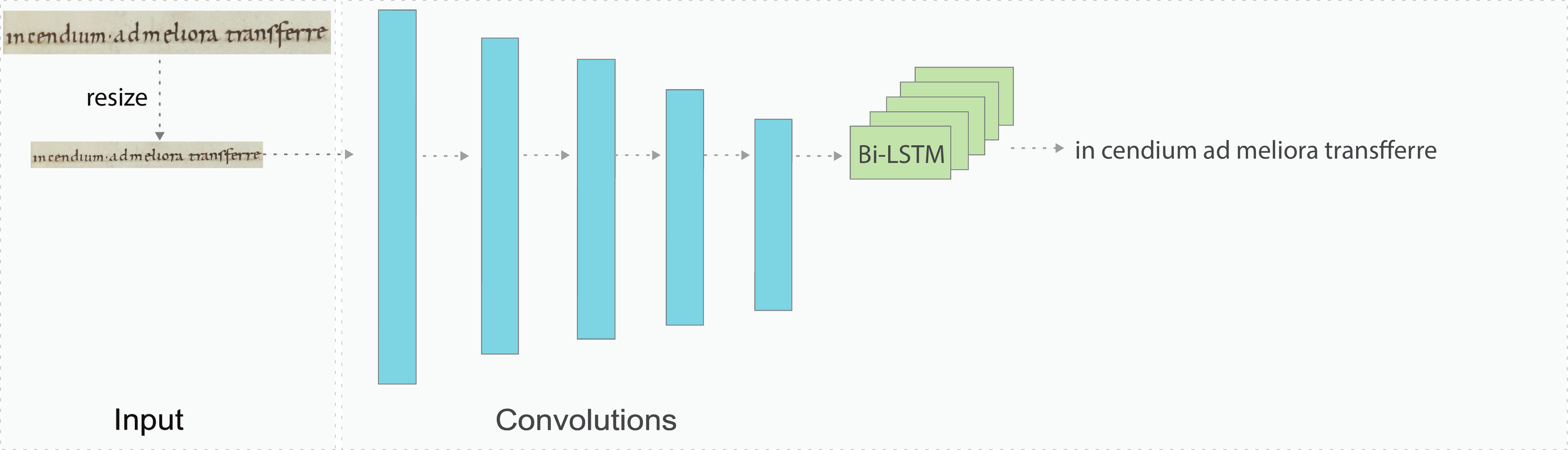}
 \caption{The complete network structure of the CRNN architecture. At the \textit{input stage}, an image of text is fed into the network at the line level. In the \textit{convolutional stage}, convolutional layers extract features from the input image. The \textit{RNN stage} extracts plaintext from the feature maps of the convolutional layers using a Bidirectional LSTM.}
 \label{fig:pipeline2}
\end{figure*}

\begin{figure}[h]
 \centering
 \includegraphics[width=\textwidth]{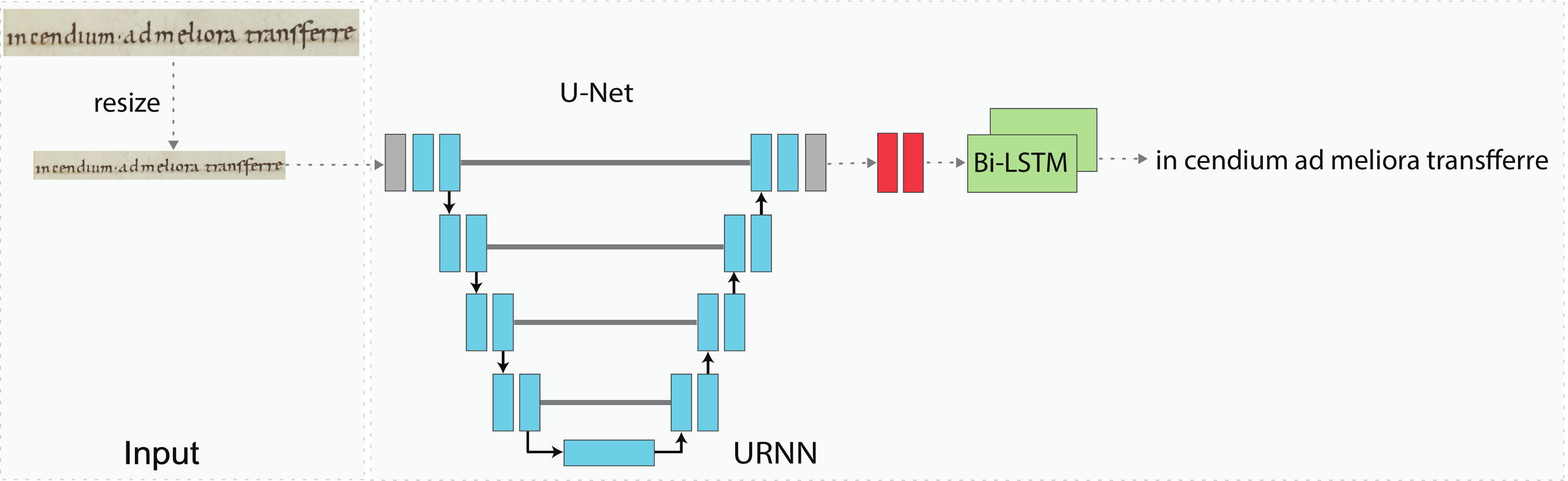}
 \caption{The complete network structure of the URNN architecture. At the \textit{input stage}, an image of text is fed into the network at the word or line level. In the \textit{U-Net stage}, characters are segmented based on visual appearance. The \textit{RNN stage} extracts plaintext from the feature maps of the U-Net using a Bidirectional LSTM.}
 \vspace{-3mm}
 \label{fig:pipeline1}
\end{figure}

\begin{figure}[ht!]
 \centering
 \includegraphics[width=\textwidth]{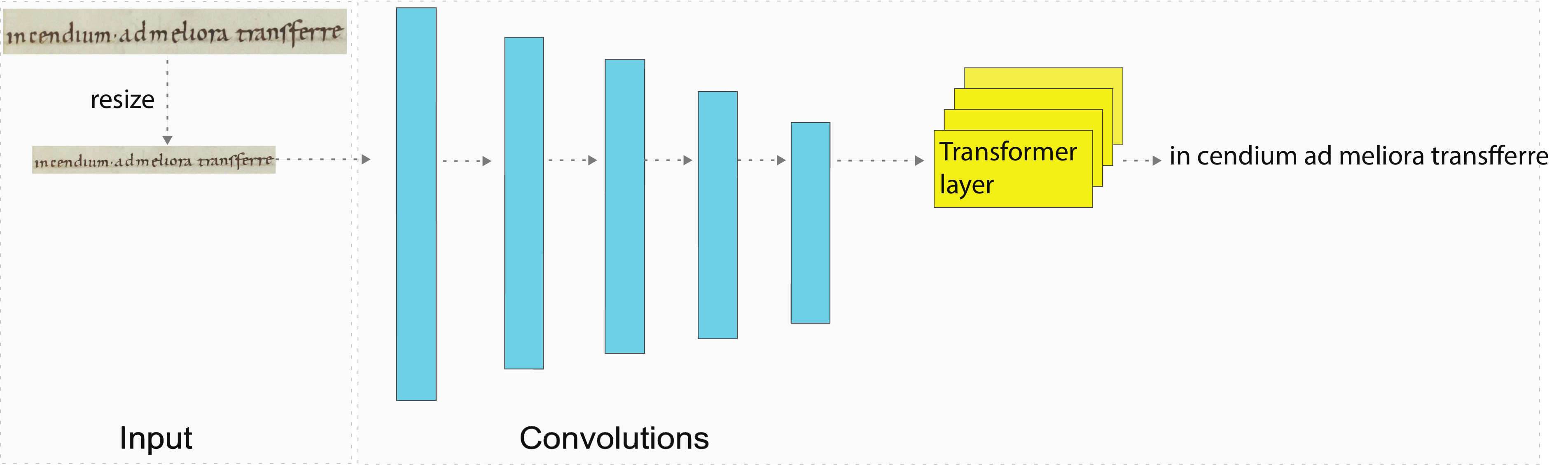}
 \caption{The complete network structure of the Ctransformer architecture. At the \textit{input stage}, an image of text is fed into the network at the word or line level. In the \textit{convolution stage}, characters are recognized based on visual appearance. The \textit{transformer stage} extracts plaintext from the feature maps of the network by using a transformer encoder with eight heads.}
 \vspace{-3mm}
 \label{fig:pipeline}
\end{figure}
\clearpage

\section{Detailed Breakdown of all Folds for Tables II \& III in Main Paper}

Here we have included the results for all of the individual experiments run for the proposed loss function prioritizing easy examples. Each row in Tables~\ref{tab:sup_crnn},~\ref{tab:sup_urnn}~and~\ref{tab:sup_ctrans} contains the experiments using the CRNN, URNN, and CTransformer models respectively. The Init.~column specifies which randomized initialization was used as a starting point. The ``V. CER'', ``V. WER'', ``T. CER'', and ``T. WER'' Columns refer to the Validation Character and Word Error Rates, and Test Character and Word Error Rates respectively. The ``Error'' row is the Standard Error of each set of experiments. The best result for each set of experiments is listed in bold.
Tables~\ref{tab:best_results_val}~and~\ref{tab:best_results_test} contain aggregations similar to the average results shown in Tables II and III of the main paper, but for the best results (highlighted in bold in Tables~\ref{tab:sup_crnn},~\ref{tab:sup_urnn}~and~\ref{tab:sup_ctrans}) on each dataset by each type of model.

\input{supp_table_crnn.tex}
\input{supp_table_urnn.tex}
\input{supp_table_ctrans.tex}
\input{best_res_val_tab.tex}
\input{best_res_tab.tex}

\clearpage

\section{Results for Putting the Priority on the Hard Examples}

The following tables correspond to Tables II and III in the main paper. Here the proposed psychophysical loss function (Eqs. 2 \& 3 in the main paper) is applied during training with the priority on hard examples (as determined by measured human reaction time), instead of the easy examples. Comparing these results to those with the priority on easy examples, we see mixed performance. In some cases there is is a reduction in the error rates (\textit{e.g.}, CRNN on all datasets), while in other cases the error is higher than the baseline (\textit{e.g.}, CTransformer on all datasets). The loss function priority should be treated as a hyperparameter, with a decision on the priority made after evaluating both modes using a validation set.  

\input{avg_res_rev_val_tab.tex}
\input{avg_res_rev_test_tab.tex}

\clearpage

\section{Qualitative Examples}
This section includes qualitative examples from a number of selections from the Test Sets of the IAM~\cite{marti2002iam} (Table~\ref{tab:sample_outputs_IAM}), RIMES~\cite{augustin2006rimes} (Table~\ref{tab:sample_outputs_rimes}), and Latin (Table~\ref{tab:sample_outputs_latin}) datasets. The ``Psych'' and ``No Psych'' rows include the prediction from models trained from the same initialization with the standard CTC loss and the psychophysical loss (prioritizing easy examples) respectively. The ``GT'' row shows the ground-truth annotation from the dataset.

\input{qual_IAM}
\input{qual_rimes}
\input{qual_lat}
\newpage

\section{Control Experiments}

\textbf{Supplemental Information on the Hard Example Mining Experiment in Sec.~V of the Main Paper.} Hard example mining, found in popular vision approaches like OHEM~\cite{Shrivastava_2016_CVPR} and Focal Loss~\cite{Lin_2017_ICCV}, attempts to algorithmically identify examples that are routinely misclassified, and reweights them accordingly during training to improve performance. This approach can be used as an alternative way to determine the difficulty of individual examples, for use with the proposed loss function in this paper. We performed an experiment to assess the suitability of our loss function with algorithmically identified hard examples. To do this, we took the baseline CRNN results from each fold (there are five total folds), averaged the character error rates for each example (where $m$ in Eq.~2 of the Main Paper is the largest character error rate), and used those averages as the associated penalties for the examples. For a fair comparison against the psychophysical loss function using psychophysical measurements, approximately 35\% of the data were considered to carry additional penalties for the loss function. In this case, the fraction that was most difficult. Training then proceeded using the psychophysical loss function in the same manner as the experiments reported for the psychophysical data, with priority placed on the least difficult of the hard examples (\textit{i.e.}, characters that are difficult, but likely not illegible). The major difference from the original experiment is that we are emphasizing the same examples the networks specifically have trouble with in training, as opposed to the examples humans have trouble with. A potential advantage of using hard examples instead of psychophysical annotations is that it is much easier to algorithmically calculate penalties, compared to running behavioral experiments with people. 

The results in Table~IV of the Main Paper (in the row labeled `M') show that by using hard examples in conjunction with the proposed psychophysical loss function, results that are better than those from the baseline CRNN (in the row labeled `B') can be achieved. However, next to the new loss function with psychophysical measurements (`E' \& `H'), the results are comparable (\textit{i.e.}, within the error of `E' and close to `H') for CER, and a bit worse for WER. This experiment demonstrates that the proposed loss function is useful for data other than psychophysical measurements. 

\textbf{Supplemental Information on the Curriculum Learning Experiment in Sec.~V of the Main Paper.} This approach, as introduced by Bengio et al.~\cite{bengio2009curriculum}, was specifically designed with artificial neural networks in mind. This makes it a good candidate as a comparison to our proposed loss function for training. Further, it also presents an opportunity to see if we can make use of psychophysical measurements in a different training setting.

We performed an experiment with the CRNN architecture that defined a curriculum by taking all of the available psychophysical measurements for the IAM dataset, sorting them by reaction time per character, and then dividing them into five groups based on this ordering of difficulty. The first group was twice as large as the others, with 875 examples (to provide a sufficient initial basis for training), and was shown right away at the beginning of training. Each subsequent group, containing 437 examples, was introduced after performance on the validation set did not improve for 20 epochs. Groups were added cumulatively during training. This experiment is an alternative way to use the psychophysical measurements outside of the loss function we propose.

The results in Table~IV of the Main Paper (in the row labeled `C') show improvement over the baseline CRNN results (in the row labeled `B'). This confirms that the psychophysical measurements help machine learning training without the need of a specialized loss function. Compared to using the same data with the psychophysical loss function with easy examples prioritized (in the row labeled `E'), this curriculum training leads to a lower character error rate on the test set, but a worse word error rate on that same set. Compared to the psychophysical loss function with hard examples prioritized (in the row labeled `H'), it results in slightly higher errors for the test set. The results from both approaches are close, however. In summary, we see curriculum learning with psychophysical data to be a promising future direction of work, including the formulation of new curricula, as well as new compatible loss functions.



\textbf{Supplemental Information on the Experiment Using Text Length in Place of Psychophysical Measurements with the Proposed Loss Function in Sec.~V of the Main Paper.} While we show that psychophysical measurements lead to good performance improvement with the proposed loss function, the loss function does not require that we use just those measurements as annotations. To assess another measure of example difficulty, we replaced the psychophysical measurements for IAM with the text length of each example (shorter  assumed  to  be  easier). Because it is trivial to collect this type of data (CTC loss requires the text length as a parameter, thus it was readily available), we did this across all of the examples. We then trained using the psychophysical loss as we did for the experiments reported for the psychophysical data. 

The results in Table~IV of the Main Paper (in the row labeled `L') showed some improvement over the baseline CRNN (`B') --- but not significant improvement. Even with having annotations for all of the data, these results were not nearly as good as those using psychophysical measurements (`E' and `H'), or a hard example mining approach (`M'). 

\section{Incorporation of a Language Model and Synthetic Pre-Training}

 The language model described in Sec.~V of the Main Paper was trained on the Wellington~\cite{wellington2}, LoB~\cite{johansson1978manual} and Brown Corpora~\cite{francis1979brown}, with the sections of the LoB Corpus that make up the Test and Validation sets of IAM removed. We used an n-gram forecasting language model with a word beam search~\cite{scheidl2018wordbeamsearch} to decode the raw predictions, with smoothing set to $0.01$, and a beam size of $60$.

In addition to the language model, we also find that by increasing the image size to $128$ pixels, which is approximately the average height in the dataset, and adding synthetic pre-training, the results become much closer to the state-of-the-art on IAM. The synthetic pre-training was done by generating $186,488$ line images from the corpus used to train the language model. We generated synthetic data using the same methodology as Xiao et al.~\cite{xiao2019deep}, albeit their model used $1,000,000$ lines of text for synthetic data generation. 

{\small
\bibliographystyle{IEEEtran}
\bibliography{egbib}
}

%% file: other_models.tex
\begin{table}[t!]
\centering
\noindent
\small\addtolength{\tabcolsep}{-2pt}
\begin{tabular}{c|c|c|c|c|}
\cline{2-5}
                                & \multicolumn{4}{c|}{Performance on Latin Dataset from Sec.~\ref{sec:expsetup}}                                               \\ \cline{2-5} 
                                & \multicolumn{2}{c|}{Validation Set}                      & \multicolumn{2}{c|}{Test Set}                      \\ \cline{2-5} 
                                & CER                           & \multicolumn{1}{c|}{WER} & \multicolumn{1}{c|}{CER} & \multicolumn{1}{c|}{WER} \\ \hline
\multicolumn{1}{|c|}{Tesseract~\cite{smith2007overview}} & \multicolumn{1}{r|}{49.65\%}  & 105.40\%                 & 44.42\%                  & 104.80\%                 \\ \hline
\multicolumn{1}{|c|}{Ocropus~\cite{breuel2008ocropus}}   & \multicolumn{1}{r|}{53.02\%}  & 113.17\%                 & 53.02\%                  & 113.17\%                 \\ \hline
\multicolumn{1}{|c|}{FineReader~\cite{FineReader}}    & \multicolumn{1}{r|}{66.11\%} &       98.48\%          & 70.19\%                  & 97.25\%                 \\ \hline
\multicolumn{1}{|c|}{Ocular~\cite{ocular}}    & \multicolumn{1}{r|}{103.84\%} & 114.23\%                 & 96.32\%                  & 115.64\%                 \\ \hline
\end{tabular}
\vspace{1mm}
\caption{Examples of off-the-shelf tools that could be used for a handwritten Latin transcription task. Character (CER) and word (WER) error rates above $100\%$ can occur when the predicted strings are longer than the correct solution, causing more insertion errors than the length of the string.}
\label{tab:other_baselines}
\vspace{-8mm}
\end{table}

%% file: avg_res_val_tab.tex
\begin{table*}[]
\begin{tabular}{c|c|c|c|c||c|c|c|c||c|c||c|c|}
\cline{2-13}
                            & \multicolumn{12}{c|}{Average Results on Validation Sets}                                                                                                    \\ \cline{2-13} 
                            & \multicolumn{12}{c|}{CRNN}                                                                                                                                  \\ \cline{2-13} 
                            & \multicolumn{4}{c||}{Without Psychophysical Loss}  & \multicolumn{4}{c||}{With Psychophysical Loss  (Easy Priority)}     & \multicolumn{2}{c||}{ \% Imp.} & \multicolumn{2}{c|}{Raw Imp.}              \\ \cline{2-13} 
                            & CER (\%)     & ERR (\%)    & WER (\%)     & ERR (\%)    & CER (\%)     & ERR (\%)    & WER (\%)     & ERR (\%)    & CER (\%)          & WER (\%)          & \multicolumn{1}{c|}{CE} & \multicolumn{1}{c|}{WE} \\ \hline
\multicolumn{1}{|c|}{I}   & 6.38  & $\pm$0.11 & 20.60 & $\pm$0.31 & 6.16  & $\pm$0.13 & 19.96 & $\pm$0.32 & 3.39       & 3.04       & 111.4                   & 40.2                    \\ \hline
\multicolumn{1}{|c|}{R} & 7.29  & $\pm$0.36 & 18.61 & $\pm$0.95 & 6.13  & $\pm$0.16 & 15.41 & $\pm$0.58 & 15.52      & 16.89      & 997.4                   & 526.4                   \\ \hline
\multicolumn{1}{|c|}{L} & 8.65  & $\pm$0.07 & 36.69 & $\pm$0.18 & 8.37  & $\pm$0.05 & 36.25 & $\pm$0.39 & 3.17       & 1.19       & 17.6                    & 3.8                     \\ \hline
                            & \multicolumn{12}{c|}{URNN}                                                                                                                                  \\ \cline{2-13} 
                            & \multicolumn{4}{c||}{Without Psychophysical Loss}  & \multicolumn{4}{c||}{With Psychophysical Loss  (Easy Priority)}     & \multicolumn{2}{c||}{ \% Imp.} & \multicolumn{2}{c|}{Raw Imp.}              \\ \cline{2-13} 
                            & CER (\%)     & ERR (\%)    & WER (\%)     & ERR (\%)    & CER (\%)     & ERR (\%)    & WER (\%)     & ERR (\%)    & CER (\%)          & WER (\%)          & \multicolumn{1}{c|}{CE} & \multicolumn{1}{c|}{WE} \\ \hline
\multicolumn{1}{|c|}{I}   & 9.20  & $\pm$3.59 & 28.91 & $\pm$9.26 & 8.74  & $\pm$3.23 & 28.04 & $\pm$8.56 & 3.48       & 2.18       & 155.3                   & 68.3                    \\ \hline
\multicolumn{1}{|c|}{R} & 6.66  & $\pm$0.31 & 18.55   & $\pm$1.11 & 6.53  & $\pm$0.45 & 17.94 & $\pm$1.53 & 2.05       & 3.58       & 44.3                    & 43                      \\ \hline
\multicolumn{1}{|c|}{L} & 12.37 & $\pm$0.55 & 49.16  & $\pm$2.01 & 11.64 & $\pm$0.81 & 47.88 & $\pm$2.95 & 6.08       & 2.77       & 60                      & 16.3                    \\ \hline
                            & \multicolumn{12}{c|}{CTransformer}                                                                                                                          \\ \cline{2-13} 
                            & \multicolumn{4}{c||}{Without Psychophysical Loss}  & \multicolumn{4}{c||}{With Psychophysical Loss  (Easy Priority)}     & \multicolumn{2}{c||}{ \% Imp.} & \multicolumn{2}{c|}{Raw Imp.}              \\ \cline{2-13} 
                            & CER (\%)    & ERR (\%)    & WER (\%)     & ERR (\%)    & CER (\%)     & ERR (\%)    & WER (\%)     & ERR (\%)    & CER (\%)          & WER (\%)          & \multicolumn{1}{c|}{CE} & \multicolumn{1}{c|}{WE} \\ \hline
\multicolumn{1}{|c|}{I}   & 10.52 & $\pm$0.19 & 34.86 & $\pm$0.51 & 10.27 & $\pm$0.12 & 34.47 & $\pm$0.32 & 2.33       & 1.07       & 17.6                    & 3.8                     \\ \hline
\multicolumn{1}{|c|}{R} & 7.41  & $\pm$0.16 & 21.06 & $\pm$0.55 & 7.07  & $\pm$0.10 & 18.86 & $\pm$0.87 & 4.53       & 10.08      & 187.6                   & 443                     \\ \hline
\multicolumn{1}{|c|}{L} & 12.13 & $\pm$0.15 & 54.93 & $\pm$0.33 & 11.88 & $\pm$0.12 & 55.39 & $\pm$0.92 & 2.05       & -0.85      & 10                      & -7.6                    \\ \hline
\end{tabular}

\vspace{2mm}

\caption{A comparison of the average results on the Validation Set of each dataset for each set of experiments. ``I" stands for IAM, ``R" for RIMES, and ``L'' for the Latin dataset. ``Raw Improvement'' refers to the number of characters or words that performance improved by when using the psychophysical loss. ``ERR'' is standard error. Improvement is observed in all but one case.}
\label{tab:average_results_val}
\vspace{-3mm}
\end{table*}

%% file: avg_res_test_tab.tex
\begin{table*}[]
\begin{tabular}{c|c|c|c|c||c|c|c|c||c|c||c|c|}
\cline{2-13}
                            & \multicolumn{12}{c|}{Average Results on Test Sets}                                                                                                    \\ \cline{2-13} 
                            & \multicolumn{12}{c|}{CRNN}                                                                                                                                  \\ \cline{2-13} 
                            & \multicolumn{4}{c||}{Without Psychophysical Loss}  & \multicolumn{4}{c||}{With Psychophysical Loss (Easy Priority)}     & \multicolumn{2}{c||}{ \% Imp.} & \multicolumn{2}{c|}{Raw Imp.}              \\ \cline{2-13} 
                            & CER (\%)     & ERR (\%)    & WER (\%)     & ERR (\%)    & CER (\%)     & ERR (\%)    & WER (\%)     & ERR (\%)    & CER (\%)          & WER (\%)          & \multicolumn{1}{c|}{CE} & \multicolumn{1}{c|}{WE} \\ \hline
\multicolumn{1}{|c|}{I} & 8.00  & $\pm$0.10 & 24.49 & $\pm$0.29 & 7.76  & $\pm$0.14 & 23.70 & $\pm$0.37 & 2.95       & 3.19       & 445.6                   & 285.6                   \\ \hline
\multicolumn{1}{|c|}{R} & 7.47  & $\pm$0.36 & 19.28 & $\pm$0.95 & 6.33  & $\pm$0.15 & 16.15 & $\pm$0.67 & 16.37      & 16.04      & 961.8                   & 558.6                   \\ \hline
\multicolumn{1}{|c|}{L} & 6.78  & $\pm$0.06 & 32.11 & $\pm$0.29 & 6.45  & $\pm$0.04 & 30.80 & $\pm$0.31 & 4.77       & 4.06       & 40.4                    & 25.2                    \\ \hline
                            & \multicolumn{12}{c|}{URNN}                                                                                                                                  \\ \cline{2-13} 
                            & \multicolumn{4}{c||}{Without Pyschophysical Loss}  & \multicolumn{4}{c||}{With Pyschophysical Loss (Easy Priority)}     & \multicolumn{2}{c||}{ \% Imp.} & \multicolumn{2}{c|}{Raw Imp.}              \\ \cline{2-13} 
                            & CER (\%)     & ERR (\%)    & WER (\%)     & ERR (\%)    & CER (\%)     & ERR (\%)    & WER (\%)     & ERR (\%)    & CER (\%)          & WER (\%)          & \multicolumn{1}{c|}{CE} & \multicolumn{1}{c|}{WE} \\ \hline
\multicolumn{1}{|c|}{I}   & 11.28 & $\pm$4.09 & 33.03 & $\pm$9.60 & 10.93 & $\pm$3.80 & 32.48 & $\pm$8.99 & 2.25       & 1.00       & 93                      & 109                     \\ \hline
\multicolumn{1}{|c|}{R} & 6.75  & $\pm$0.31 & 18.86 & $\pm$1.00 & 6.66  & $\pm$0.47 & 18.36 & $\pm$1.61 & 1.56       & 3.03       & 485.7                   & 182.7                   \\ \hline
\multicolumn{1}{|c|}{L} & 8.98  & $\pm$0.29 & 41.18 & $\pm$1.11 & 8.59  & $\pm$0.40 & 40.07 & $\pm$1.74 & 4.40       & 2.76       & 60.3                    & 25.3                    \\ \hline
                            & \multicolumn{12}{c|}{CTransformer}                                                                                                                          \\ \cline{2-13} 
                            & \multicolumn{4}{c||}{Without Pyschophysical Loss}  & \multicolumn{4}{c||}{With Pyschophysical Loss  (Easy Priority)}     & \multicolumn{2}{c||}{ \% Imp.} & \multicolumn{2}{c|}{Raw Imp.}              \\ \cline{2-13} 
                            & CER (\%)    & ERR (\%)    & WER (\%)     & ERR (\%)    & CER (\%)     & ERR (\%)    & WER (\%)     & ERR (\%)    & CER (\%)          & WER (\%)          & \multicolumn{1}{c|}{CE} & \multicolumn{1}{c|}{WE} \\ \hline
\multicolumn{1}{|c|}{I}   & 13.06 & $\pm$0.21 & 39.93 & $\pm$0.47 & 12.79 & $\pm$0.11 & 39.49 & $\pm$0.28 & 1.98       &  1.05      &  461                   &  164.4                   \\ \hline
\multicolumn{1}{|c|}{R} & 7.58  & $\pm$0.18 & 21.68 & $\pm$0.62 & 7.27  & $\pm$0.12 & 20.62 & $\pm$0.48 & 4.01       &  4.76      &  309.8                 &  200                     \\ \hline
\multicolumn{1}{|c|}{L} & 9.81  & $\pm$0.16 & 49.59 & $\pm$0.68 & 8.96  & $\pm$0.07 & 47.17 & $\pm$0.48 & 8.57       &  4.82      &  99.6                  &  46.8                    \\ \hline
\end{tabular}

\vspace{2mm}

\vspace{2mm}

\caption{A comparison of the average results on the Test Set of each dataset for each set of experiments. All cases improve.}
\label{tab:average_results_test}
\vspace{-5mm}
\end{table*}

%% file: table_control.tex
\begin{table}[t]
\noindent
\centering
\small\addtolength{\tabcolsep}{-1pt}
\begin{tabular}{l|l|l|l|l|l|l|l|l|}
\cline{2-9}
                                      
                                         & \multicolumn{4}{c|}{IAM Validation (\%)} & \multicolumn{4}{c|}{IAM Test (\%)} \\ \cline{2-9} 
                                         & CER   &  ERR    & WER    & ERR   & CER  & ERR      & WER     & ERR  \\ \hline
\multicolumn{1}{|l|}{B} & 6.38   &  $\pm$.11  & 20.60      & $\pm$.31 & 8.00 & $\pm$.10    & 24.49  & $\pm$.29    \\ \hline
\multicolumn{1}{|l|}{E} & 6.16 &  $\pm$.13    & 19.96     & $\pm$.32 & 7.76 &  $\pm$.14   & 23.70 & $\pm$.37     \\ \hline
\multicolumn{1}{|l|}{H} & 6.00 & $\pm$.03    & 19.62 & $\pm$.08 & 7.56 & $\pm$.02  &  23.41 & $\pm$.03       \\ \hline \hline
\multicolumn{1}{|l|}{M} & 6.16  &   $\pm$.06  &   20.33    & $\pm$.19  &  7.76  & $\pm$.07 & 24.02  &  $\pm$.21   \\ \hline
\multicolumn{1}{|l|}{C} & 5.73 &  $\pm$.10   & 19.11     & $\pm$.27 & 7.68 & $\pm$.11   & 23.92 &  $\pm$.29   \\ \hline
\multicolumn{1}{|l|}{L} & 6.28 &  $\pm$.11   & 20.33    & $\pm$.21 & 7.92 & $\pm$.08  &  24.21 & $\pm$.19      \\ \hline
\end{tabular}
\vspace{1mm}
\caption{Average results for the control experiments on the IAM dataset using the CRNN architecture.}
\vspace{-9mm}
\label{tab:control}
\end{table}

%% file: table_comparison.tex
\begin{table}[b]
\noindent
\centering
\small\addtolength{\tabcolsep}{-1pt}
\begin{tabular}{l|l|l|l|l|}
\cline{2-5}
                                      
                                         & \multicolumn{2}{l|}{IAM Val. (\%)} & \multicolumn{2}{l|}{IAM Test (\%)} \\ \cline{2-5} 
                                         & CER         & WER        & CER         & WER        \\ \hline
\multicolumn{1}{|l|}{Voigtlander et al.~\cite{voigtlaender2016handwriting}} & 2.4       & 7.1      & 3.5       & 9.3       \\ \hline
\multicolumn{1}{|l|}{Puigcerver et al.~\cite{puigcerver2017multidimensional}} & 2.9       & 9.2      & 4.4       & 12.2        \\ \hline
\multicolumn{1}{|l|}{Xiao et al.~\cite{xiao2019deep}}       & 1.96      & 6.63     & 3.03      & 8.66      \\ \hline \hline
\multicolumn{1}{|l|}{CRNN w/o Psych. Loss}             &    3.00          & 8.72           &      4.71      &    12.6       \\ \hline
\multicolumn{1}{|l|}{CRNN w/ Psych. Loss (E)}             &    2.79          & 8.19           & 4.35           & 11.09          \\ \hline
\multicolumn{1}{|l|}{CRNN w/ Psych. Loss (H)}               & 2.88           & 8.29          & 4.21           & 10.68          \\ \hline
\end{tabular}
\vspace{1mm}
\caption{Comparison to other work on handwritten document transcription with language model post-correction. %
}
\vspace{-8mm}
\label{tab:comparison}
\end{table}

%% file: table1_examples_of_lines.tex
\begin{table*}[h]
\noindent
\centering
\small\addtolength{\tabcolsep}{-1pt}
\begin{tabular}{|l|c|c|c|c|}
\hline

Dataset
    & Easier for Humans &$r$
    & Harder for Humans &$r$
    \\\hline
IAM
    &\includegraphics[width=.37\textwidth]{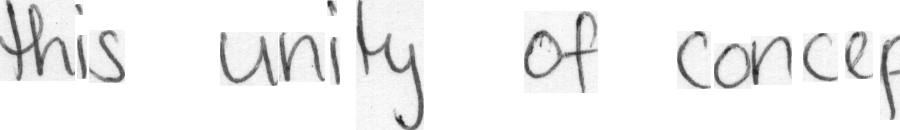}&0.162
    &\includegraphics[width=.37\textwidth]{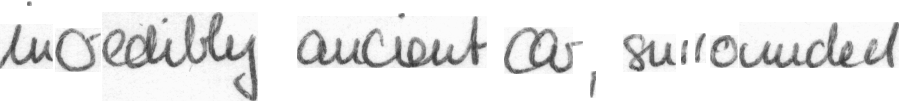}&30.99
    \\\hline
RIMES
    &\includegraphics[width=.37\textwidth]{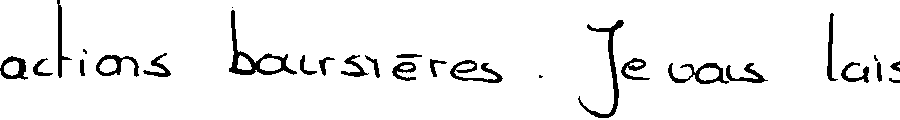}&0.072
    &\includegraphics[width=.37\textwidth]{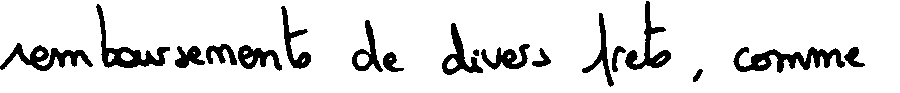}&51.06
\\\hline

Latin    
    &\includegraphics[width=.37\textwidth]{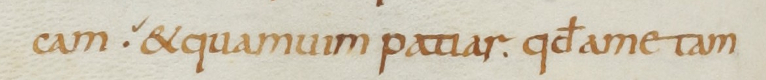}&2.655
    &\includegraphics[width=.37\textwidth]{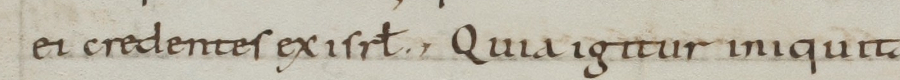}&218.7
    \\\hline

\end{tabular}
\vspace{1mm}
\caption{Examples of lines from each dataset that our human annotators found easy and difficult. The $r$ columns contain the average reaction times associated with the lines in milliseconds.}
\label{tab:dataset_ex}
\end{table*}

%% file: supp_table_crnn.tex
\begin{table*}[!htbp]
\noindent
\centering
\small\addtolength{\tabcolsep}{-1pt}
\begin{tabular}{l|r|r|r|r||r|r|r|r||r|r|r|r|}
\cline{2-13}
\multicolumn{1}{c|}{}         & \multicolumn{12}{c|}{IAM Aachen Split}                                                                                                                                                                                                                                                                                                                                                \\ \cline{2-13} 
                              & \multicolumn{4}{c||}{Without Psychophysical Loss (\%)}                                                                         & \multicolumn{4}{c||}{With Psychophysical Loss (\%)}                                                                            & \multicolumn{4}{c|}{Improvement (\%)}                                                                                     \\ \hline
\multicolumn{1}{|l|}{Init.}    & \multicolumn{1}{l|}{V. CER}  & \multicolumn{1}{l|}{V. WER}  & \multicolumn{1}{l|}{T. CER}   & \multicolumn{1}{l||}{T. WER}  & \multicolumn{1}{l|}{V. CER}  & \multicolumn{1}{l|}{V. WER}  & \multicolumn{1}{l|}{T. CER}   & \multicolumn{1}{l||}{T. WER}  & \multicolumn{1}{l|}{V. CER}  & \multicolumn{1}{l|}{V. WER}  & \multicolumn{1}{l|}{T. CER}   & \multicolumn{1}{l|}{T. WER}   \\ \hline
\multicolumn{1}{|l|}{1}       & \textbf{6.10}                & \textbf{19.75}               & \textbf{7.73}                 & \textbf{23.54}                & 6.28                         & 20.26                        & 7.79                          & 23.72                         & -2.85                        & -2.57                        & -0.80                         & -0.77                         \\ \hline
\multicolumn{1}{|l|}{2}       & 6.40                         & 20.63                        & 8.20                          & 24.91                         & 6.15                         & 20.31                        & 7.71                          & 23.69                         & 3.92                         & 1.54                         & 5.99                          & 4.91                          \\ \hline
\multicolumn{1}{|l|}{3}       & 6.79                         & 21.70                        & 8.24                          & 25.17                         & 6.57                         & 20.79                        & 8.27                          & 25.03                         & 3.20                         & 4.19                         & -0.37                         & 0.55                          \\ \hline
\multicolumn{1}{|l|}{4}       & 6.37                         & 20.52                        & 8.04                          & 24.67                         & 5\textbf{.86}                & \textbf{19.05}               & \textbf{7.48}                 & \textbf{22.83}                & 8.00                         & 7.15                         & 6.92                          & 7.48                          \\ \hline
\multicolumn{1}{|l|}{5}       & 6.25                         & 20.38                        & 7.81                          & 24.13                         & 5.95                         & 19.38                        & 7.57                          & 23.23                         & 4.70                         & 4.88                         & 3.01                          & 3.75                          \\ \hline
\multicolumn{1}{|l|}{Avg.}    & 6.38                         & 20.60                        & 8.00                          & 24.49                         & 6.16                         & 19.96                        & 7.76                          & 23.70                         & 3.39                         & 3.04                         & 2.95                          & 3.19                          \\ \hline
\multicolumn{1}{|l|}{Error}   & $\pm$0.11                    & $\pm$0.31                    & $\pm$0.10                     & $\pm$0.29                     & $\pm$0.13                    & $\pm$0.32                    & $\pm$0.14                     & $\pm$0.37                     & \multicolumn{4}{l}{}                                                                                                        \\ \hline
\multicolumn{1}{c|}{}         & \multicolumn{12}{c|}{Raw IAM Aachen Split*}                                                                                                                                                                                                                                                                                                                                             \\ \cline{2-13} 
                              & \multicolumn{4}{c||}{Without Psychophysical Loss (\%)}                                                                         & \multicolumn{4}{c||}{With Psychophysical Loss (\%)}                                                                            & \multicolumn{4}{c|}{Improvement (\%)}                                                                                     \\ \hline
\multicolumn{1}{|l|}{Init.}      & \multicolumn{1}{l|}{V. CER}  & \multicolumn{1}{l|}{V. WER}  & \multicolumn{1}{l|}{T. CER}   & \multicolumn{1}{l||}{T. WER}  & \multicolumn{1}{l|}{V. CER}  & \multicolumn{1}{l|}{V. WER}  & \multicolumn{1}{l|}{T. CER}   & \multicolumn{1}{l||}{T. WER}  & \multicolumn{1}{l|}{V. CER}  & \multicolumn{1}{l|}{V. WER}  & \multicolumn{1}{l|}{T. CER}   & \multicolumn{1}{l|}{T. WER}   \\ \hline
\multicolumn{1}{|l|}{1}       & 7.77                         & 21.92                        & 11.83                         & 25.37                         & 6.98                         & 20.92                        & 11.01                         & \textbf{24.14}                & 10.23                        & 4.57                         & 6.87                          & 4.85                          \\ \hline
\multicolumn{1}{|l|}{2}       & \textbf{7.26}                & \textbf{20.78}               & \textbf{11.10}                & \textbf{24.15}                & 6.73                         & 21.11                        & \textbf{10.17}                & 24.76                         & 7.27                         & -1.56                        & 8.35                          & -2.53                         \\ \hline
\multicolumn{1}{|l|}{3}       & 7.35                         & 21.41                        & 11.60                         & 25.11                         & 7.28                         & 20.42                        & 11.17                         & 24.23                         & 0.91                         & 4.65                         & 3.69                          & 3.50                          \\ \hline
\multicolumn{1}{|l|}{4}       & 8.11                         & 22.74                        & 12.30                         & 26.40                         & \textbf{6.40}                & 21.05                        & 10.82                         & 24.32                         & 21.16                        & 7.40                         & 12.03                         & 7.89                          \\ \hline
\multicolumn{1}{|l|}{5}       & 8.02                         & 22.64                        & 12.28                         & 26.48                         & 7.03                         & \textbf{20.40}               & 11.04                         & 23.83                         & 12.25                        & 9.90                         & 10.13                         & 10.00                         \\ \hline
\multicolumn{1}{|l|}{Avg.}    & 7.70                         & 21.90                        & 11.82                         & 25.50                         & 6.88                         & 20.78                        & 10.84                         & 24.25                         & 10.36                        & 4.99                         & 8.21                          & 4.74                          \\ \hline
\multicolumn{1}{|l|}{Error}   & $\pm$0.17                    & $\pm$0.37                    & $\pm$0.23                     & $\pm$0.43                     & $\pm$0.15                    & $\pm$0.15                    & $\pm$0.18                     & $\pm$0.15                     & \multicolumn{4}{l}{}                                                                                                        \\ \hline
\multicolumn{1}{c|}{}         & \multicolumn{12}{c|}{RIMES}                                                                                                                                                                                                                                                                                                                                                             \\ \cline{2-13} 
                              & \multicolumn{4}{c||}{Without Psychophysical Loss (\%)}                                                                        & \multicolumn{4}{l||}{With Psychophysical Loss (\%)}                                                                            & \multicolumn{4}{c|}{Improvement (\%)}                                                                                      \\ \hline
\multicolumn{1}{|l|}{Init.}    & \multicolumn{1}{l|}{V. CER}  & \multicolumn{1}{l|}{V. WER}  & \multicolumn{1}{l|}{T. CER}   & \multicolumn{1}{l||}{T. WER}  & \multicolumn{1}{l|}{V. CER}  & \multicolumn{1}{l|}{V. WER}  & \multicolumn{1}{l|}{T. CER}   & \multicolumn{1}{l||}{T. WER}  & \multicolumn{1}{l|}{V. CER}  & \multicolumn{1}{l|}{V. WER}  & \multicolumn{1}{l|}{T. CER}   & \multicolumn{1}{l|}{T. WER}   \\ \hline
\multicolumn{1}{|l|}{1}       & 7.56                         & 19.64                        & 7.80                          & 20.31                         & 6.23                         & 15.82                        & 6.53                          & 16.84                         & 17.56                        & 19.47                        & 16.3                          & 17.10                         \\ \hline
\multicolumn{1}{|l|}{2}       & \textbf{6.37}                & \textbf{15.78}               & \textbf{6.47}                 & \textbf{16.39}                & \textbf{5.79}                & 14.39                        & 5.94                          & 14.81                         & 9.11                         & 8.80                         & 8.26                          & 9.64                          \\ \hline
\multicolumn{1}{|l|}{3}       & 6.95                         & 17.79                        & 7.20                          & 18.47                         & 5.83                         & \textbf{14.15}               & \textbf{5.93}                 & \textbf{14.77}                & 16.09                        & 20.46                        & 17.57                         & 20.02                         \\ \hline
\multicolumn{1}{|l|}{4}       & 7.05                         & 18.35                        & 7.23                          & 19.10                         & 6.10                         & 15.28                        & 6.31                          & 15.99                         & 13.42                        & 16.77                        & 12.72                         & 16.30                         \\ \hline
\multicolumn{1}{|l|}{5}       & 8.51                         & 21.47                        & 8.64                          & 22.12                         & 6.69                         & 17.40                        & 6.93                          & 18.33                         & 21.40                        & 18.95                        & 26.97                         & 17.14                         \\ \hline
\multicolumn{1}{|l|}{Avg.}    & 7.29                         & 18.61                        & 7.47                          & 19.28                         & 6.13                         & 15.41                        & 6.33                          & 16.15                         & 15.52                        & 16.89                        & 16.37                         & 16.04                         \\ \hline
\multicolumn{1}{|l|}{Error}   & $\pm$0.36                    & $\pm$0.95                    & $\pm$0.36                     & $\pm$0.95                     & $\pm$0.16                    & $\pm$0.58                    & $\pm$0.15                     & $\pm$0.67                     & \multicolumn{4}{l}{}                                                                                                        \\ \hline
\multicolumn{1}{c|}{}         & \multicolumn{12}{c|}{Latin}                                                                                                                                                                                                                                                                                                                                                             \\ \cline{2-13} 
                              & \multicolumn{4}{c||}{Without Psychophysical Loss (\%)}                                                                         & \multicolumn{4}{c||}{With Psychophysical Loss (\%)}                                                                            & \multicolumn{4}{c|}{Improvement (\%)}                                                                                     \\ \hline
\multicolumn{1}{|l|}{Init.}      & \multicolumn{1}{l|}{V. CER}  & \multicolumn{1}{l|}{V. WER}  & \multicolumn{1}{l|}{T. CER}   & \multicolumn{1}{l||}{T. WER}  & \multicolumn{1}{l|}{V. CER}  & \multicolumn{1}{l|}{V. WER}  & \multicolumn{1}{l|}{T. CER}   & \multicolumn{1}{l||}{T. WER}  & \multicolumn{1}{l|}{V. CER}  & \multicolumn{1}{l|}{V. WER}  & \multicolumn{1}{l|}{T. CER}   & \multicolumn{1}{l|}{T. WER}   \\ \hline
\multicolumn{1}{|l|}{1}       & 8.56                         & 36.44                        & 6.77                          & 31.99                         & 8.51                         & 37.35                        & 6.51                          & 31.53                         & 0.62                         & -2.50                        & 3.81                          & 1.45                          \\ \hline
\multicolumn{1}{|l|}{2}       & 8.60                         & \textbf{36.37}               & 6.88                          & 32.27                         & 8.42                         & 36.96                        & 6.52                          & 31.56                         & 2.09                         & -1.63                        & 5.19                          & 2.20                          \\ \hline
\multicolumn{1}{|l|}{3}       & 8.85                         & 37.18                        & 6.91                          & 32.98                         & 8.40                         & 35.97                        & 6.51                          & 30.47                         & 5.06                         & 3.26                         & 5.82                          & 7.60                          \\ \hline
\multicolumn{1}{|l|}{4}       & \textbf{8.46}                & 36.40                        & \textbf{6.57}                 & \textbf{31.19}                & 8.27                         & \textbf{35.32}               & \textbf{6.33}                 & 30.23                         & 2.17                         & 2.97                         & 3.67                          & 3.06                          \\ \hline
\multicolumn{1}{|l|}{5}       & 8.77                         & 37.07                        & 6.75                          & 32.12                         & \textbf{8.25}                & 35.65                        & 6.39                          & \textbf{30.20}                & 5.91                         & 3.85                         & 5.36                          & 5.98                          \\ \hline
\multicolumn{1}{|l|}{Avg.}    & 8.65                         & 36.69                        & 6.78                          & 32.11                         & 8.37                         & 36.25                        & 6.45                          & 30.80                         & 3.17                         & 1.19                         & 4.77                          & 4.06                          \\ \hline
\multicolumn{1}{|l|}{Error}   & $\pm$0.07                    & $\pm$0.18                    & $\pm$0.06                     & $\pm$0.29                     & $\pm$0.05                    & $\pm$0.39                    & $\pm$0.04                     & $\pm$0.31                          & \multicolumn{4}{l}{}                                                                                                   \\ \cline{1-9}

\end{tabular}
\vspace{1mm}
\caption{All results for experiments using the CRNN architecture.\newline\newline*This set includes results for an alternate version of the Aachen split. We found that the numbers of lines in the published datasets do not line up with the Aachen split as published. This is because there is often spurious text from the forms that IAM was originally written on, found in the last lines of each page. We corrected for this, and presented results from this more commonly used version in the main paper. However, since we had run experiments on the original split using CRNN, we include them here.}
\label{tab:sup_crnn}
\end{table*}

%% file: supp_table_urnn.tex
\begin{table*}[!htbp]
\noindent
\centering
\small\addtolength{\tabcolsep}{-1pt}
\begin{tabular}{l|r|r|r|r||r|r|r|r||r|r|r|r|}
\cline{2-13}
\multicolumn{1}{c|}{}         & \multicolumn{12}{c|}{IAM Aachen Split}                                                                                                                                                                                                                                                                                                                                                \\ \cline{2-13} 
                              & \multicolumn{4}{c||}{Without Psychophysical Loss (\%)}                                                                         & \multicolumn{4}{c||}{With Psychophysical Loss (\%)}                                                                            & \multicolumn{4}{c|}{Improvement (\%)}                                                                                     \\ \hline
\multicolumn{1}{|l|}{Init.}    & \multicolumn{1}{l|}{V. CER}  & \multicolumn{1}{l|}{V. WER}  & \multicolumn{1}{l|}{T. CER}   & \multicolumn{1}{l||}{T. WER}  & \multicolumn{1}{l|}{V. CER}  & \multicolumn{1}{l|}{V. WER}  & \multicolumn{1}{l|}{T. CER}   & \multicolumn{1}{l||}{T. WER}  & \multicolumn{1}{l|}{V. CER}  & \multicolumn{1}{l|}{V. WER}  & \multicolumn{1}{l|}{T. CER}   & \multicolumn{1}{l|}{T. WER}   \\ \hline
\multicolumn{1}{|l|}{1}       & \textbf{5.30}                & \textbf{19.08}               & \textbf{6.99}                 & \textbf{22.91}                & \textbf{5.36}                & \textbf{18.97}                & \textbf{6.92}                 & \textbf{22.71}                & -1.12                        & 0.55                         & 0.98                         & 0.87                          \\ \hline
\multicolumn{1}{|l|}{2}       & 5.92                         & 20.24                        & 7.41                          & 23.98                         & 5.67                         & 20.00                         & 7.33                          & 24.28                         & 4.32                         & 1.21                         & 1.00                         & -1.27                          \\ \hline
\multicolumn{1}{|l|}{3}       & 16.37                        & 47.42                        & 19.45                         & 52.22                         & 15.19                        & 45.16                         & 18.52                         & 50.45                         & 7.24                         & 4.77                         & 4.76                         & 3.39                          \\ \hline
\multicolumn{1}{|l|}{Avg.}    & 9.20                         & 28.91                        & 11.28                         & 33.03                         & 8.74                         & 28.04                         & 10.93                         & 32.48                         & 3.48                         & 2.18                         & 2.25                         & 1.00                          \\ \hline
\multicolumn{1}{|l|}{Error}   & $\pm$3.59                    & $\pm$9.26                    & $\pm$4.09                     & $\pm$9.60                     & $\pm$3.23                    & $\pm$8.56                     & $\pm$3.80                     & $\pm$8.99                     & \multicolumn{4}{l}{}                                                                                                        \\ \hline
\multicolumn{1}{c|}{}         & \multicolumn{12}{c|}{RIMES}                                                                                                                                                                                                                                                                                                                                                             \\ \cline{2-13} 
                              & \multicolumn{4}{c||}{Without Psychophysical Loss (\%)}                                                                        & \multicolumn{4}{l||}{With Psychophysical Loss (\%)}                                                                            & \multicolumn{4}{c|}{Improvement (\%)}                                                                                      \\ \hline
\multicolumn{1}{|l|}{Init.}    & \multicolumn{1}{l|}{V. CER}  & \multicolumn{1}{l|}{V. WER}  & \multicolumn{1}{l|}{T. CER}   & \multicolumn{1}{l||}{T. WER}  & \multicolumn{1}{l|}{V. CER}  & \multicolumn{1}{l|}{V. WER}  & \multicolumn{1}{l|}{T. CER}   & \multicolumn{1}{l||}{T. WER}  & \multicolumn{1}{l|}{V. CER}  & \multicolumn{1}{l|}{V. WER}  & \multicolumn{1}{l|}{T. CER}   & \multicolumn{1}{l|}{T. WER}   \\ \hline
\multicolumn{1}{|l|}{1}       & \textbf{6.35}                & \textbf{17.15}                & \textbf{6.43}                & \textbf{17.77}                & \textbf{5.88}                & \textbf{15.75}                & \textbf{6.03}                & \textbf{16.15}                & 7.38                         & 8.19                         & 6.21                         & 9.12                          \\ \hline
\multicolumn{1}{|l|}{2}       & 7.27                         & 20.73                         & 7.37                         & 20.87                         & 7.39                         & 20.89                         & 7.56                         & 21.50                         & -1.54                         & -0.78                       & -2.63                        & -3.03                         \\ \hline
\multicolumn{1}{|l|}{3}       & 6.35                         & 17.77                         & 6.45                         & 17.95                         & 6.33                         & 17.18                         & 6.38                         & 17.41                         & 0.30                         & 3.35                         & 1.09                         & 2.99                          \\ \hline
\multicolumn{1}{|l|}{Avg.}    & 6.66                         & 18.55                         & 6.75                         & 18.86                         & 6.53                         & 17.94                         & 6.66                         & 18.36                         & 2.05                         & 3.58                         & 1.56                         & 3.03                          \\ \hline
\multicolumn{1}{|l|}{Error}   & $\pm$0.31                    & $\pm$1.11                     & $\pm$0.31                    & $\pm$1.00                     & $\pm$0.45                    & $\pm$1.53                    & $\pm$0.47                   & $\pm$1.61                     & \multicolumn{4}{l}{}                                                                                                          \\ \hline
\multicolumn{1}{c|}{}         & \multicolumn{12}{c|}{Latin}                                                                                                                                                                                                                                                                                                                                                             \\ \cline{2-13} 
                              & \multicolumn{4}{c||}{Without Psychophysical Loss (\%)}                                                                         & \multicolumn{4}{c||}{With Psychophysical Loss (\%)}                                                                            & \multicolumn{4}{c|}{Improvement (\%)}                                                                                     \\ \hline
\multicolumn{1}{|l|}{Init.}      & \multicolumn{1}{l|}{V. CER}  & \multicolumn{1}{l|}{V. WER}  & \multicolumn{1}{l|}{T. CER}   & \multicolumn{1}{l||}{T. WER}  & \multicolumn{1}{l|}{V. CER}  & \multicolumn{1}{l|}{V. WER}  & \multicolumn{1}{l|}{T. CER}   & \multicolumn{1}{l||}{T. WER}  & \multicolumn{1}{l|}{V. CER}  & \multicolumn{1}{l|}{V. WER}  & \multicolumn{1}{l|}{T. CER}   & \multicolumn{1}{l|}{T. WER}   \\ \hline
\multicolumn{1}{|l|}{1}       & 11.91                         & 47.24                         & 8.89                         & 41.21                         & \textbf{10.76}                & \textbf{44.20}                & 8.28                         & 39.11                         & 9.65                         & 6.44                         & 6.91                         & 5.10                          \\ \hline
\multicolumn{1}{|l|}{2}       & 13.47                         & 53.19                         & 9.53                         & 43.10                         & 13.26                         & 53.72                         & 9.38                         & 43.44                         & 1.54                         & -1.00                         & 1.54                         & -0.81                         \\ \hline
\multicolumn{1}{|l|}{3}       & \textbf{11.73}                & \textbf{47.06}                & \textbf{8.52}                & \textbf{39.24}                & 10.90                         & 45.71                         & \textbf{8.12}                & \textbf{37.67}                & 7.04                         & 2.87                         & 4.75                         & 3.99                          \\ \hline
\multicolumn{1}{|l|}{Avg.}    & 12.37                         & 49.16                         & 8.98                         & 41.18                         & 11.64                         & 47.88                         & 8.59                         & 40.07                         & 6.08                         & 2.77                         & 4.40                         & 2.76                          \\ \hline
\multicolumn{1}{|l|}{Error}   & $\pm$0.55                     & $\pm$2.01                     & $\pm$0.29                    & $\pm$1.11                     & $\pm$0.81                     & $\pm$2.95                     & $\pm$0.40                    & $\pm$1.74                     & \multicolumn{4}{l}{}  \\\cline{1-9}                                                        

\end{tabular}
\vspace{1mm}
\caption{All results for experiments using the URNN architecture.}
\label{tab:sup_urnn}
\end{table*}

%% file: supp_table_ctrans.tex
\begin{table*}[!htbp]
\noindent
\centering
\small\addtolength{\tabcolsep}{-1pt}
\begin{tabular}{l|r|r|r|r||r|r|r|r||r|r|r|r|}
\cline{2-13}
\multicolumn{1}{c|}{}         & \multicolumn{12}{c|}{IAM Aachen Split}                                                                                                                                                                                                                                                                                                                                                \\ \cline{2-13} 
                              & \multicolumn{4}{c||}{Without Psychophysical Loss (\%)}                                                                         & \multicolumn{4}{c||}{With Psychophysical Loss (\%)}                                                                            & \multicolumn{4}{c|}{Improvement (\%)}                                                                                     \\ \hline
\multicolumn{1}{|l|}{Init.}    & \multicolumn{1}{l|}{V. CER}  & \multicolumn{1}{l|}{V. WER}  & \multicolumn{1}{l|}{T. CER}   & \multicolumn{1}{l||}{T. WER}  & \multicolumn{1}{l|}{V. CER}  & \multicolumn{1}{l|}{V. WER}  & \multicolumn{1}{l|}{T. CER}   & \multicolumn{1}{l||}{T. WER}  & \multicolumn{1}{l|}{V. CER}  & \multicolumn{1}{l|}{V. WER}  & \multicolumn{1}{l|}{T. CER}   & \multicolumn{1}{l|}{T. WER}   \\ \hline
\multicolumn{1}{|l|}{1}       & 10.52                         & 34.63                         & 13.17                         & 40.44                         & 10.30                         & 34.46                         & 12.90                         & 39.76                         & 2.09                         & 0.48                         & 2.05                         & 1.68                         \\ \hline
\multicolumn{1}{|l|}{2}       & \textbf{10.04}                & \textbf{33.77}                & 12.66                         & 38.92                         & \textbf{9.82}                 & \textbf{33.25}                & \textbf{12.37}                & \textbf{38.44}                & 2.21                         & 1.53                         & 2.34                         & 1.25                          \\ \hline
\multicolumn{1}{|l|}{3}       & 10.18                         & 33.86                         & \textbf{12.56}                & \textbf{38.86}                & 10.58                         & 35.06                         & 13.04                         & 39.97                         & -3.99                         & -3.54                         & -3.78                         & -2.87                          \\ \hline
\multicolumn{1}{|l|}{4}       & 11.10                         & 36.34                         & 13.72                         & 41.36                         & 10.28                         & 34.60                         & 12.83                         & 39.36                         & 7.37                         & 4.80                         & 6.44                         & 4.83                          \\ \hline
\multicolumn{1}{|l|}{5}       & 10.78                         & 35.71                         & 13.18                         & 40.07                         & 10.36                         & 34.96                         & 12.81                         & 39.92                         & 3.96                         & 2.10                         & 2.83                         & 0.36                          \\ \hline
\multicolumn{1}{|l|}{Avg.}    & 10.52                         & 34.86                         & 13.06                         & 39.93                         & 10.27                         & 34.47                         & 12.79                         & 39.49                         & 2.33                         & 1.07                         & 1.98                         & 1.05                          \\ \hline
\multicolumn{1}{|l|}{Error}   & $\pm$0.19                    & $\pm$0.51                    & $\pm$0.21                     & $\pm$0.47                     & $\pm$0.12                    & $\pm$0.32                    & $\pm$0.11                     & $\pm$0.28                     & \multicolumn{4}{l}{}                                                                                                        \\ \hline
\multicolumn{1}{c|}{}         & \multicolumn{12}{c|}{RIMES}                                                                                                                                                                                                                                                                                                                                                             \\ \cline{2-13} 
                              & \multicolumn{4}{c||}{Without Psychophysical Loss (\%)}                                                                        & \multicolumn{4}{l||}{With Psychophysical Loss (\%)}                                                                            & \multicolumn{4}{c|}{Improvement (\%)}                                                                                      \\ \hline
\multicolumn{1}{|l|}{Init.}    & \multicolumn{1}{l|}{V. CER}  & \multicolumn{1}{l|}{V. WER}  & \multicolumn{1}{l|}{T. CER}   & \multicolumn{1}{l||}{T. WER}  & \multicolumn{1}{l|}{V. CER}  & \multicolumn{1}{l|}{V. WER}  & \multicolumn{1}{l|}{T. CER}   & \multicolumn{1}{l||}{T. WER}  & \multicolumn{1}{l|}{V. CER}  & \multicolumn{1}{l|}{V. WER}  & \multicolumn{1}{l|}{T. CER}   & \multicolumn{1}{l|}{T. WER}   \\ \hline
\multicolumn{1}{|l|}{1}       & 7.66                         & 21.51                         & 7.77                         & 21.92                         & 7.20                         & 20.73                         & 7.47                         & 21.92                         & 6.01                         & 3.63                         & 3.92                         & 0.03                         \\ \hline
\multicolumn{1}{|l|}{2}       & \textbf{6.84}                & \textbf{19.29}                & \textbf{6.95}                & \textbf{19.68}                & \textbf{6.81}                & 19.19                         & 6.93                         & 19.53                         & 0.44                         & 0.52                         & 0.39                         & 0.77                          \\ \hline
\multicolumn{1}{|l|}{3}       & 7.72                         & 22.41                         & 7.99                         & 23.23                         & 7.30                         & \textbf{15.68}                & 7.49                         & 21.09                         & 5.44                         & 30.03                         & 6.17                         & 9.23                          \\ \hline
\multicolumn{1}{|l|}{4}       & 7.25                         & 20.36                         & 7.41                         & 21.02                         & 6.86                         & 18.62                         & 7.05                         & 19.50                         & 5.38                         & 8.55                         & 4.85                         & 7.20                          \\ \hline
\multicolumn{1}{|l|}{5}       & 7.59                         & 21.74                         & 7.79                         & 22.54                         & 7.18                         & 20.07                         & 7.42                         & 21.06                         & 5.40                         & 7.68                         & 4.73                         & 6.57                          \\ \hline
\multicolumn{1}{|l|}{Avg.}    & 7.41                         & 21.06                         & 7.58                         & 21.68                         & 7.07                         & 18.86                         & 7.27                         & 20.62                         & 4.53                         & 10.08                         & 4.01                         & 4.76                          \\ \hline
\multicolumn{1}{|l|}{Error}   & $\pm$0.16                    & $\pm$0.55                    & $\pm$0.18                     & $\pm$0.62                     & $\pm$0.10                    & $\pm$0.87                    & $\pm$0.12                     & $\pm$0.48                     & \multicolumn{4}{l}{}                                                                                                        \\  \hline
\multicolumn{1}{c|}{}         & \multicolumn{12}{c|}{Latin}                                                                                                                                                                                                                                                                                                                                                             \\ \cline{2-13} 
                              & \multicolumn{4}{c||}{Without Psychophysical Loss (\%)}                                                                         & \multicolumn{4}{c||}{With Psychophysical Loss (\%)}                                                                            & \multicolumn{4}{c|}{Improvement (\%)}                                                                                     \\ \hline
\multicolumn{1}{|l|}{Init.}      & \multicolumn{1}{l|}{V. CER}  & \multicolumn{1}{l|}{V. WER}  & \multicolumn{1}{l|}{T. CER}   & \multicolumn{1}{l||}{T. WER}  & \multicolumn{1}{l|}{V. CER}  & \multicolumn{1}{l|}{V. WER}  & \multicolumn{1}{l|}{T. CER}   & \multicolumn{1}{l||}{T. WER}  & \multicolumn{1}{l|}{V. CER}  & \multicolumn{1}{l|}{V. WER}  & \multicolumn{1}{l|}{T. CER}   & \multicolumn{1}{l|}{T. WER}   \\ \hline
\multicolumn{1}{|l|}{1}       & 12.29                         & 55.80                         & 10.39                        & 52.05                         & \textbf{11.62}                & 56.52                         & \textbf{8.82}                & 47.33                         & 5.46                         & -1.29                         & 15.12                         & 9.06                         \\ \hline
\multicolumn{1}{|l|}{2}       & 12.05                         & 54.42                         & 9.86                         & 49.82                         & 11.86                         & 53.70                         & 8.88                         & 46.25                         & 1.57                         & 1.33                         & 9.93                         & 7.16                          \\ \hline
\multicolumn{1}{|l|}{3}       & 12.54                         & 55.51                         & 9.69                         & 48.34                         & 12.21                         & 56.24                         & 9.18                         & 47.66                         & 2.69                         & -1.31                         & 5.25                         & 1.42                          \\ \hline
\multicolumn{1}{|l|}{4}       & 12.15                         & 54.91                         & \textbf{9.52}                & \textbf{48.30}                & 11.63                         & \textbf{52.79}                & 8.88                         & \textbf{45.97}                & 4.28                         & 3.86                         & 6.66                         & 4.81                         \\ \hline
\multicolumn{1}{|l|}{5}       & \textbf{11.63}                & \textbf{54.01}                & 9.61                         & 49.44                         & 12.07                         & 57.71                         & 9.04                         & 48.63                         & -3.73                         & -6.86                         & 5.88                         & 1.64                          \\ \hline
\multicolumn{1}{|l|}{Avg.}    & 12.13                         & 54.93                         & 9.81                         & 49.59                         & 11.88                         & 55.39                         & 8.96                         & 47.17                         & 2.05                         & -0.85                         & 8.57                         & 4.82                          \\ \hline
\multicolumn{1}{|l|}{Error}   & $\pm$0.15                    & $\pm$0.33                    & $\pm$0.16                     & $\pm$0.68                     & $\pm$0.12                    & $\pm$0.92                    & $\pm$0.07                     & $\pm$0.48                     & \multicolumn{4}{l}{}                                                                                                        \\  \cline{1-9}

\end{tabular}
\vspace{1mm}
\caption{All results for experiments using the CTransformer architecture.}
\label{tab:sup_ctrans}
\end{table*}

%% file: best_res_val_tab.tex
\begin{table}[h]
\centering
\begin{tabular}{c|c|c||c|c||c|c|}
\cline{2-7}
                            & \multicolumn{6}{c|}{Best Results on Validation Sets}                                                                                                                      \\ \cline{2-7} 
                            & \multicolumn{6}{c|}{CRNN}                                                                                                                                           \\ \cline{2-7} 
                            & \multicolumn{2}{c||}{Without Psychophysical Loss}                      & \multicolumn{2}{c||}{With Psychophysical Loss}                     & \multicolumn{2}{c||}{\% Improvement}                         \\ \cline{2-7} 
                            & CER (\%)                          & \multicolumn{1}{c||}{WER (\%)} & \multicolumn{1}{c|}{CER (\%)} & \multicolumn{1}{c||}{WER (\%)} & \multicolumn{1}{c|}{CER} & \multicolumn{1}{c||}{WER} \\ \hline
\multicolumn{1}{|c|}{IAM}   & \multicolumn{1}{c|}{6.10}  &   19.75                &   5.86                 & 19.05                  &  3.89                  &  3.54                  \\ \hline
\multicolumn{1}{|c|}{RIMES} & \multicolumn{1}{c|}{6.37}  &   15.78                &   5.79                 & 14.15                  &  9.11                  & 10.34                  \\ \hline
\multicolumn{1}{|c|}{LATIN} & \multicolumn{1}{c|}{8.46}  &   36.37                &   8.25                 & 35.32                  &  2.39                  &  2.87                  \\ \hline
                            & \multicolumn{6}{c|}{URNN}                                                                                                                                           \\ \cline{2-7} 
                            & \multicolumn{2}{c||}{Without Psychophysical Loss}                      & \multicolumn{2}{c||}{With Psychophysical Loss}                     & \multicolumn{2}{c||}{\% Improvement}                         \\ \cline{2-7} 
                            & CER (\%)                          & \multicolumn{1}{c||}{WER (\%)} & \multicolumn{1}{c|}{CER (\%)} & \multicolumn{1}{c||}{WER (\%)} & \multicolumn{1}{c|}{CER} & \multicolumn{1}{c||}{WER} \\ \hline
\multicolumn{1}{|c|}{IAM}   & \multicolumn{1}{c|}{5.30}  & 19.08                   & 5.36                   & 18.97                  & -1.12                   & 0.55                   \\ \hline
\multicolumn{1}{|c|}{RIMES} & \multicolumn{1}{c|}{6.35}  & 17.15                  & 5.88                   & 15.75                  &  7.33                  &  8.19                  \\ \hline
\multicolumn{1}{|c|}{LATIN} & \multicolumn{1}{c|}{11.73}  & 47.06                  & 10.76                   & 44.20                  & 8.24                   & 6.08                   \\ \hline
                            & \multicolumn{6}{c|}{CTransformer}                                                                                                                                   \\ \cline{2-7} 
                            & \multicolumn{2}{c||}{Without Psychophysical Loss}                      & \multicolumn{2}{c||}{With Psychophysical Loss}                     & \multicolumn{2}{c||}{\% Improvement}                         \\ \cline{2-7} 
                            & CER (\%)                          & \multicolumn{1}{c||}{WER (\%)} & \multicolumn{1}{c|}{CER (\%)} & \multicolumn{1}{c||}{WER (\%)} & \multicolumn{1}{c|}{CER} & \multicolumn{1}{c||}{WER} \\ \hline
\multicolumn{1}{|c|}{IAM}   & \multicolumn{1}{c|}{10.04} & 33.77                  & 9.82                  & 33.25                  & 2.21                   & 1.53                   \\ \hline
\multicolumn{1}{|c|}{RIMES} & \multicolumn{1}{c|}{6.84} &  19.29                 &  6.81                 &  15.68                 & 0.44                   &  18.71                 \\ \hline
\multicolumn{1}{|c|}{LATIN} & \multicolumn{1}{c|}{11.63} & 54.01                  & 11.62                  &  52.79                 & 0.12                   &  2.25                 \\ \hline
\end{tabular}
\vspace{1mm}
\caption{A comparison of the models from each set of experiments with the lowest validation set CER.}
\label{tab:best_results_val}
\vspace{-3mm}
\end{table}

%% file: best_res_tab.tex
\begin{table}[h]
\centering
\begin{tabular}{c|c|c||c|c||c|c|}
\cline{2-7}
                            & \multicolumn{6}{c|}{Best Results on Test Sets}                                                                                                                      \\ \cline{2-7} 
                            & \multicolumn{6}{c|}{CRNN}                                                                                                                                           \\ \cline{2-7} 
                            & \multicolumn{2}{c||}{Without Psychophysical Loss}                      & \multicolumn{2}{c||}{With Psychophysical Loss}                     & \multicolumn{2}{c||}{\% Improvement}                         \\ \cline{2-7} 
                            & CER (\%)                          & \multicolumn{1}{c||}{WER (\%)} & \multicolumn{1}{c|}{CER (\%)} & \multicolumn{1}{c||}{WER (\%)} & \multicolumn{1}{c|}{CER} & \multicolumn{1}{c||}{WER} \\ \hline
\multicolumn{1}{|c|}{IAM}   & \multicolumn{1}{c|}{7.73}  & 23.54                  & 7.48                   & 22.83                  & 3.19                   & 3.04                   \\ \hline
\multicolumn{1}{|c|}{RIMES} & \multicolumn{1}{c|}{6.47}  & 16.39                  & 5.93                   & 14.77                  & 8.29                   & 9.90                   \\ \hline
\multicolumn{1}{|c|}{LATIN} & \multicolumn{1}{c|}{6.57}  & 31.19                  & 6.33                   & 30.20                  & 3.67                   & 3.15                   \\ \hline
                            & \multicolumn{6}{c|}{URNN}                                                                                                                                           \\ \cline{2-7} 
                            & \multicolumn{2}{c||}{Without Psychophysical Loss}                      & \multicolumn{2}{c||}{With Psychophysical Loss}                     & \multicolumn{2}{c||}{\% Improvement}                         \\ \cline{2-7} 
                            & CER (\%)                          & \multicolumn{1}{c||}{WER (\%)} & \multicolumn{1}{c|}{CER (\%)} & \multicolumn{1}{c||}{WER (\%)} & \multicolumn{1}{c|}{CER} & \multicolumn{1}{c||}{WER} \\ \hline
\multicolumn{1}{|c|}{IAM}   & \multicolumn{1}{c|}{6.99}  & 22.91                  & 6.92                   & 22.71                  & 0.98                   & 0.87                   \\ \hline
\multicolumn{1}{|c|}{RIMES} & \multicolumn{1}{c|}{6.43}  & 17.77                  & 6.03                   & 16.15                  & 6.21                   & 9.12                   \\ \hline
\multicolumn{1}{|c|}{LATIN} & \multicolumn{1}{c|}{8.52}  & 39.24                  & 8.12                   & 37.67                  & 4.75                   & 3.99                   \\ \hline
                            & \multicolumn{6}{c|}{CTransformer}                                                                                                                                   \\ \cline{2-7} 
                            & \multicolumn{2}{c||}{Without Psychophysical Loss}                      & \multicolumn{2}{c||}{With Psychophysical Loss}                     & \multicolumn{2}{c||}{\% Improvement}                         \\ \cline{2-7} 
                            & CER (\%)                          & \multicolumn{1}{c||}{WER (\%)} & \multicolumn{1}{c|}{CER (\%)} & \multicolumn{1}{c||}{WER (\%)} & \multicolumn{1}{c|}{CER} & \multicolumn{1}{c||}{WER} \\ \hline
\multicolumn{1}{|c|}{IAM}   & \multicolumn{1}{c|}{12.56} & 38.86                  & 12.37                  & 38.44                  & 1.53                   & 1.08                   \\ \hline
\multicolumn{1}{|c|}{RIMES} & \multicolumn{1}{c|}{6.95}  & 19.68                  & 6.93                   & 19.50                  & 0.39                   & 0.89                   \\ \hline

\multicolumn{1}{|c|}{LATIN} & \multicolumn{1}{c|}{9.52} & 48.30                  & 8.82                  & 45.97                  & 7.30                  & 4.81                  \\ \hline
\end{tabular}
\vspace{1mm}
\caption{A comparison of the models from each set of experiments with the lowest test set CER.}
\label{tab:best_results_test}
\vspace{-3mm}
\end{table}

%% file: avg_res_rev_val_tab.tex
\begin{table*}[h]
\noindent
\centering
\small\addtolength{\tabcolsep}{-1pt}
\begin{tabular}{c|c|c|c|c||c|c|c|c||c|c||c|c|}
\cline{2-13}
                            & \multicolumn{12}{c|}{Average Results on Validation Sets}                                                                                                    \\ \cline{2-13} 
                            & \multicolumn{12}{c|}{CRNN}                                                                                                                                  \\ \cline{2-13} 
                            & \multicolumn{4}{c||}{Without Psychophysical Loss}  & \multicolumn{4}{c||}{With Psychophysical Loss  (Hard Priority)}     & \multicolumn{2}{c||}{ \% Imp.} & \multicolumn{2}{c|}{Raw Imp.}              \\ \cline{2-13} 
                            & CER (\%)     & ERR (\%)    & WER (\%)     & ERR (\%)    & CER (\%)     & ERR (\%)    & WER (\%)     & ERR (\%)    & CER (\%)          & WER (\%)          & \multicolumn{1}{c|}{CE} & \multicolumn{1}{c|}{WE} \\ \hline
\multicolumn{1}{|c|}{I}   & 6.38  & $\pm$0.11 & 20.60 & $\pm$0.31 & 6.00  & $\pm$0.03 & 19.62 & $\pm$0.08 & 5.92      &  4.65      &    296.4                &    89.2                 \\ \hline
\multicolumn{1}{|c|}{R} & 7.29  & $\pm$0.36 & 18.61 & $\pm$0.95 &  5.64  & $\pm$0.10 & 13.82 & $\pm$0.25  & 22.06      &  25.17     &   800.2                 &    452               \\ \hline
\multicolumn{1}{|c|}{L} & 8.65  & $\pm$0.07 & 36.69 & $\pm$0.18 & 8.59  & $\pm$0.07 & 36.03 & $\pm$0.23 &  0.60      &  1.82       &    4.8                 &     4.6                 \\ \hline
                            & \multicolumn{12}{c|}{URNN}                                                                                                                                  \\ \cline{2-13} 
                            & \multicolumn{4}{c||}{Without Psychophysical Loss}  & \multicolumn{4}{c||}{With Psychophysical Loss  (Hard Priority)}     & \multicolumn{2}{c||}{ \% Imp.} & \multicolumn{2}{c|}{Raw Imp.}              \\ \cline{2-13} 
                            & CER (\%)     & ERR (\%)    & WER (\%)     & ERR (\%)    & CER (\%)     & ERR (\%)    & WER (\%)     & ERR (\%)    & CER (\%)          & WER (\%)          & \multicolumn{1}{c|}{CE} & \multicolumn{1}{c|}{WE} \\ \hline
\multicolumn{1}{|c|}{I}   & 9.20  & $\pm$3.59 & 28.91 & $\pm$9.26 & 5.62  & $\pm$0.30 & 19.69 & $\pm$0.88 & 21.76      & 19.30        &  1497.7                  & 842.7                    \\ \hline
\multicolumn{1}{|c|}{R} & 6.66  & $\pm$0.31 & 18.55   & $\pm$1.11 & 6.37  & $\pm$0.37 & 17.39 & $\pm$1.11 & 3.46       & 5.01       &  141.7                   & 99.7                      \\ \hline
\multicolumn{1}{|c|}{L} & 12.37 & $\pm$0.55 & 49.16  & $\pm$2.01 &18.85  & $\pm$6.96 & 56.79 & $\pm$9.70 & -55.67       & -16.87       &  -459.7                     & -99.7                   \\ \hline
                            & \multicolumn{12}{c|}{CTransformer}                                                                                                                          \\ \cline{2-13} 
                            & \multicolumn{4}{c||}{Without Psychophysical Loss}  & \multicolumn{4}{c||}{With Psychophysical Loss  (Hard Priority)}     & \multicolumn{2}{c||}{ \% Imp.} & \multicolumn{2}{c|}{Raw Imp.}              \\ \cline{2-13} 
                            & CER (\%)    & ERR (\%)    & WER (\%)     & ERR (\%)    & CER (\%)     & ERR (\%)    & WER (\%)     & ERR (\%)    & CER (\%)          & WER (\%)          & \multicolumn{1}{c|}{CE} & \multicolumn{1}{c|}{WE} \\ \hline
\multicolumn{1}{|c|}{I}   & 10.52 & $\pm$0.19 & 34.86 & $\pm$0.51 & 10.52 & $\pm$0.27 & 35.06 & $\pm$0.58 &  -0.09      & -0.64       &  -12.2                   & -38.6                     \\ \hline
\multicolumn{1}{|c|}{R} & 7.41  & $\pm$0.16 & 21.06 & $\pm$0.55 &7.56  & $\pm$0.05 & 21.52 & $\pm$0.16 & -2.13       &  -2.39     &   -72.4                &       -46               \\ \hline
\multicolumn{1}{|c|}{L} & 12.13 & $\pm$0.15 & 49.16 & $\pm$0.33 &12.36 & $\pm$0.09 & 56.54 & $\pm$0.61 &  -1.94      &  -2.94     &      -7.6                 &    -27.6                 \\ \hline
\end{tabular}

\vspace{2mm}

\caption{A comparison of the average results on the Validation Set of each dataset for each set of experiments. ``I" stands for IAM, ``R" for RIMES, and ``L'' for the Latin dataset. ``Raw Improvement'' refers to the number of characters or words that performance improved by when using the psychophysical loss. ``ERR'' is standard error.}
\label{tab:average_results_val_rev}
\vspace{-3mm}
\end{table*}

%% file: avg_res_rev_test_tab.tex
\begin{table*}[hb]
\noindent
\centering
\small\addtolength{\tabcolsep}{-1pt}
\begin{tabular}{c|c|c|c|c||c|c|c|c||c|c||c|c|}
\cline{2-13}
                            & \multicolumn{12}{c|}{Average Results on Test Sets}                                                                                                    \\ \cline{2-13} 
                            & \multicolumn{12}{c|}{CRNN}                                                                                                                                  \\ \cline{2-13} 
                            & \multicolumn{4}{c||}{Without Psychophysical Loss}  & \multicolumn{4}{c||}{With Psychophysical Loss (Hard Priority)}     & \multicolumn{2}{c||}{ \% Imp.} & \multicolumn{2}{c|}{Raw Imp.}              \\ \cline{2-13} 
                            & CER (\%)     & ERR (\%)    & WER (\%)     & ERR (\%)    & CER (\%)     & ERR (\%)    & WER (\%)     & ERR (\%)    & CER (\%)          & WER (\%)          & \multicolumn{1}{c|}{CE} & \multicolumn{1}{c|}{WE} \\ \hline
\multicolumn{1}{|c|}{I} & 8.00  & $\pm$0.10 & 24.49 & $\pm$0.29 & 7.56  & $\pm$0.02 & 23.41 & $\pm$0.03 & 5.50 & 4.36  &   556.4                &   271               \\ \hline
\multicolumn{1}{|c|}{R} & 7.47  & $\pm$0.36 & 19.28 & $\pm$0.95 & 5.76  & $\pm$0.10 & 14.28 & $\pm$0.25 & 22.36     & 25.40      &  1406.8                   & 793                    \\ \hline
\multicolumn{1}{|c|}{L} & 6.78  & $\pm$0.06 & 32.11 & $\pm$0.29 & 6.58  & $\pm$0.08 & 30.82 & $\pm$0.34 &  2.81    & 3.96       &   25.4                 &    22.6                \\ \hline
                            & \multicolumn{12}{c|}{URNN}                                                                                                                                  \\ \cline{2-13} 
                            & \multicolumn{4}{c||}{Without Pyschophysical Loss}  & \multicolumn{4}{c||}{With Pyschophysical Loss (Hard Priority)}     & \multicolumn{2}{c||}{ \% Imp.} & \multicolumn{2}{c|}{Raw Imp.}              \\ \cline{2-13} 
                            & CER (\%)     & ERR (\%)    & WER (\%)     & ERR (\%)    & CER (\%)     & ERR (\%)    & WER (\%)     & ERR (\%)    & CER (\%)          & WER (\%)          & \multicolumn{1}{c|}{CE} & \multicolumn{1}{c|}{WE} \\ \hline
\multicolumn{1}{|c|}{I} & 11.28 & $\pm$4.09 & 33.03 & $\pm$9.60 &7.22   & $\pm$0.37 &23.68  & $\pm$1.14 & 20.57       & 17.48       & 6830.7                    & 3321.7                    \\ \hline 
\multicolumn{1}{|c|}{R} & 6.75  & $\pm$0.31 & 18.86 & $\pm$1.00 &6.49   & $\pm$0.36 &17.88  & $\pm$1.09 & 2.94       & 4.16      &  211.3                 &  164.7                     \\ \hline
\multicolumn{1}{|c|}{L} & 8.98  & $\pm$0.29 & 41.18 & $\pm$1.11 &14.25   & $\pm$14.25 &49.23  & $\pm$9.75 & -62.12       & -20.94       &  -708.7                   &  -191                   \\ \hline
                            & \multicolumn{12}{c|}{CTransformer}                                                                                                                          \\ \cline{2-13} 
                            & \multicolumn{4}{c||}{Without Pyschophysical Loss}  & \multicolumn{4}{c||}{With Pyschophysical Loss  (Hard Priority)}     & \multicolumn{2}{c||}{ \% Imp.} & \multicolumn{2}{c|}{Raw Imp.}              \\ \cline{2-13} 
                            & CER (\%)    & ERR (\%)    & WER (\%)     & ERR (\%)    & CER (\%)     & ERR (\%)    & WER (\%)     & ERR (\%)    & CER (\%)          & WER (\%)          & \multicolumn{1}{c|}{CE} & \multicolumn{1}{c|}{WE} \\ \hline
\multicolumn{1}{|c|}{I}   & 13.06 & $\pm$0.21 & 39.93 & $\pm$0.47 &13.07  & $\pm$0.24 &  40.05 & $\pm$0.48 &  -0.09     &  -0.33      &    -5.6                   & -24                     \\ \hline 
\multicolumn{1}{|c|}{R} & 7.58  & $\pm$0.18 & 21.68 & $\pm$0.62 &7.80  & $\pm$0.04 &22.49  & $\pm$0.16 &  -3.08      &  -4.05      &     -167.2               &  -108                  \\ \hline
\multicolumn{1}{|c|}{L} & 9.81  & $\pm$0.16 & 49.59 & $\pm$0.68 &9.93   & $\pm$0.09 &  50.61& $\pm$0.81 & -1.27       & -2.15       &     -10.2               &   -26.6                   \\ \hline
\end{tabular}

\vspace{2mm}

\vspace{2mm}

\caption{A comparison of the average results on the Test Set of each dataset for each set of experiments.}
\label{tab:average_results_test}
\vspace{-5mm}
\end{table*}

%% file: qual_IAM.tex
\begin{table*}[h]
\noindent
\centering
\small\addtolength{\tabcolsep}{-1pt}
\begin{tabular}{|l|l|}
\hline

Input
    &\includegraphics[width=.8\textwidth]{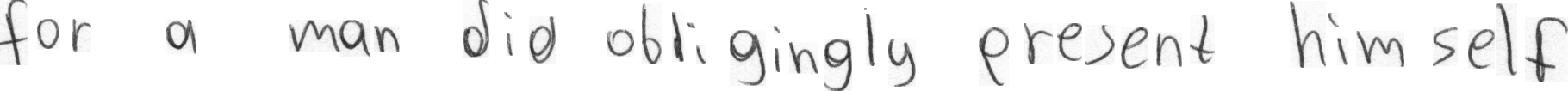}
    \\\hline
No Psych &for a man did obligingly present himself
\\\hline
Psych &for a man did obligingly present himself
\\\hline GT&for a man did obligingly present himself
\\\hline
Input
    &\includegraphics[width=.8\textwidth]{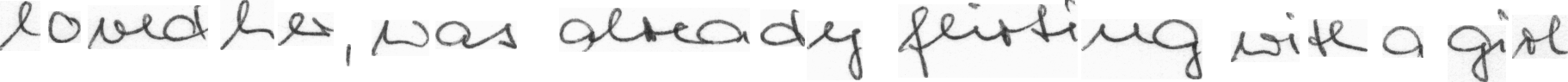}
    \\\hline
No Psych &lonedher , was already feirting vite a givl
\\\hline
Psych &loved her , was already flirting with a girl
\\\hline GT&loved her , was already flirting with a girl 
\\\hline
Input
    &\includegraphics[width=.8\textwidth]{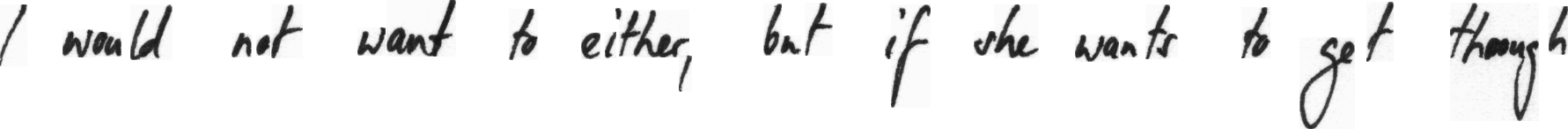}
    \\\hline
No Psych &I would not want to either , but if she wnts to get though
\\\hline
Psych &I would not want to either , but if she wants to get though
\\\hline GT&I would not want to either , but if she wants to get through
\\\hline
Input
    &\includegraphics[width=.8\textwidth]{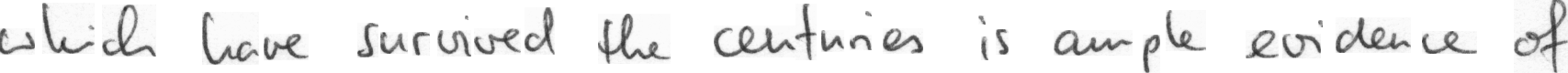}
    \\\hline
No Psych &which have survived the centrries is ample evidence of
\\\hline
Psych &which have survived the centuries is ample evidence of
\\\hline GT&which have survived the centuries is ample evidence of
\\\hline
Input
    &\includegraphics[width=.8\textwidth]{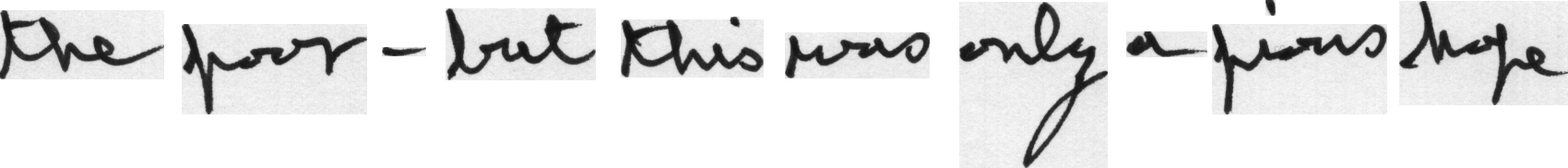}
    \\\hline
No Psych &the for-but Misturooly afinshupe
\\\hline
Psych &the paur-bue this tews only a finshage
\\\hline GT&the poor - but this was only a pious hope
\\\hline
Input
    &\includegraphics[width=.8\textwidth]{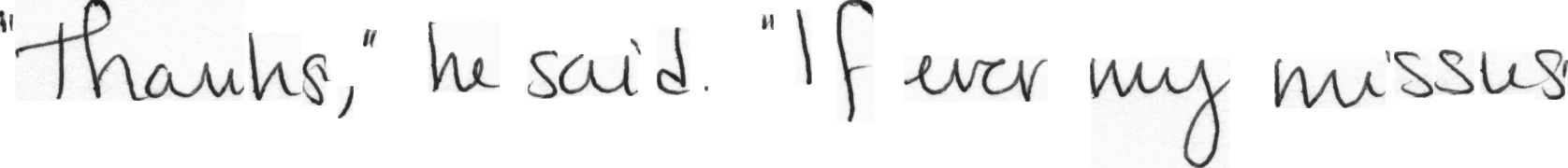}
    \\\hline
No Psych &" Thanks , he said (f evermy missus
\\\hline
Psych &" Thanks , " he said . " If ever my missus
\\\hline GT&" Thanks , " he said . " If ever my missus
\\\hline

\end{tabular}
\vspace{1mm}
\caption{Sample outputs for handwritten English lines from the IAM dataset.}
\label{tab:sample_outputs_IAM}
\end{table*}

%% file: qual_rimes.tex
\begin{table*}[htbp]
\noindent
\centering
\small\addtolength{\tabcolsep}{-1pt}
\begin{tabular}{|l|l|}
\hline

Input
    &\includegraphics[width=.8\textwidth]{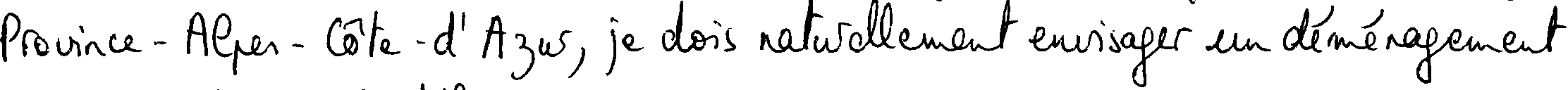}
    \\\hline
No Psych &Prvonce . Ayen Cote diAu , je dos racturellement envisayer un déménagement.
\\\hline
Psych &prvance . Ayen - Vote d' Ajour , je dois naturellement envisager un déménagement
\\\hline
GT&Province - Alpes - Côte - d' Azur , je dois naturellement envisager un déménagement
\\\hline
Input
    &\includegraphics[width=.2\textwidth]{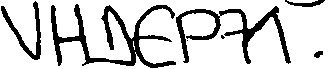}
    \\\hline
No Psych &U R D E R 7  .
\\\hline
Psych &U H D E P T R .
\\\hline GT&V H D E P 7 1 .
\\\hline
Input
    &\includegraphics[width=.8\textwidth]{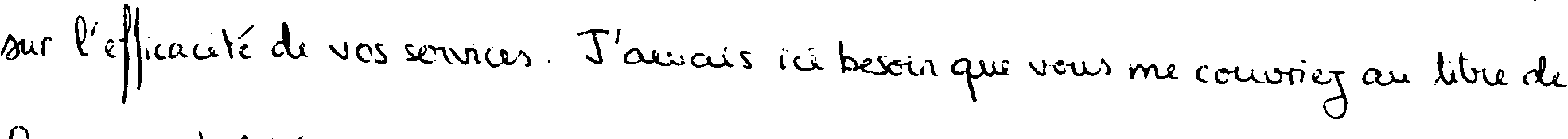}
    \\\hline
No Psych &sus l' offracité de vos services . J' avais ca beunque vous ne cousegantre de
\\\hline
Psych &sour l' offcacité de vos sonvices . J' avais i bessir que vous me courriez au titre de
\\\hline GT&sur l' efficacité de vos services . J ' aurais ici besoin que vous me couvriez au titre de
\\\hline
Input
    &\includegraphics[width=.8\textwidth]{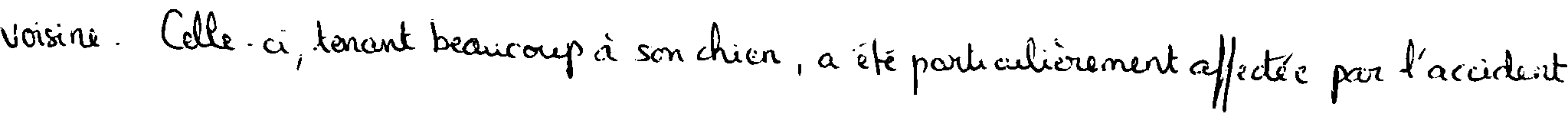}
    \\\hline
No Psych &voisine . Celle - ci - demant baucoup à sonchier , a ét portialièsement fficde par l' accident .
\\\hline
Psych &voisire . Celle - ci - tenent beaucoup à son chien , a été portialièrement affectée par l' accident
\\\hline GT&voisine . Celle - ci , tenant beaucoup à son chien , a été particulièrement affectée par l' accident
\\\hline
Input
    &\includegraphics[width=.3\textwidth]{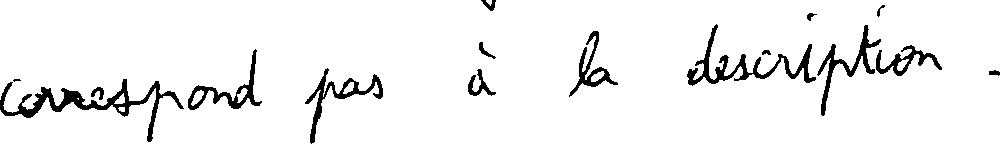}
    \\\hline
No Psych &corspond pas à la désciption .
\\\hline
Psych &correspond pas à la description .
\\\hline GT&correspond pas à la description .
\\\hline
Input
    &\includegraphics[width=.3\textwidth]{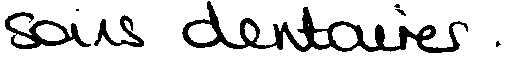}
    \\\hline
No Psych &fais dertainer .
\\\hline
Psych &sains denairers .
\\\hline GT&soins dentaires .
\\\hline

\end{tabular}
\vspace{1mm}
\caption{Sample outputs for handwritten French lines from the RIMES dataset.}
\label{tab:sample_outputs_rimes}
\end{table*}

%% file: qual_lat.tex
\begin{table*}[htbp]
\noindent
\centering
\small\addtolength{\tabcolsep}{-1pt}
\begin{tabular}{|l|l|}
\hline

Input
    &\includegraphics[width=.5\textwidth]{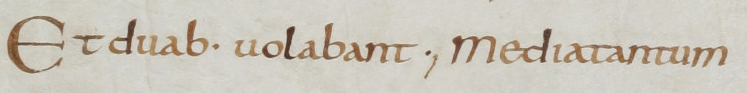}
    \\\hline
No Psych &et duab. uolabant mediatantum\\\hline
Psych & et duab. uolabant media tantum
\\\hline
GT&et duab. uolabant media tantum
\\\hline

Input
    &\includegraphics[width=.5\textwidth]{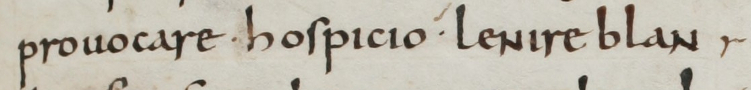}
    \\\hline
No Psych &prucare hospicio lexire blaur
\\\hline
Psych &prbouocare hospicio lenire blany 
\\\hline
GT&prouocare hospicio lenire blan
\\\hline

Input
    &\includegraphics[width=.5\textwidth]{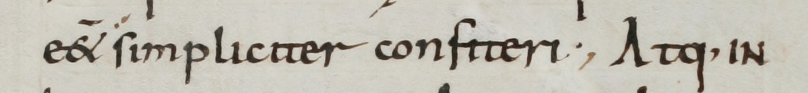}
    \\\hline
No Psych &\# simpliciter confiteri atq in\\\hline
Psych & e\#\_ simpliciter confiteri atq. in
\\\hline
GT&e\#\_ simpliciter confiteri; atq. in
\\\hline

Input
    &\includegraphics[width=.5\textwidth]{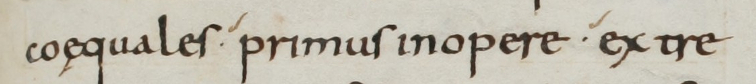}
    \\\hline
No Psych &coquales primus inopere extre\\\hline
Psych &coequales primus in opere extre 
\\\hline
GT&coe'quales primus in opere ex tre
\\\hline

Input
    &\includegraphics[width=.5\textwidth]{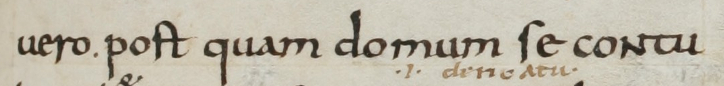}
    \\\hline
No Psych &uero postquam domum se cortu\\\hline
Psych &uero post quam domum se contu
\\\hline
GT&uero post quam domum se contu
\\\hline

Input
    &\includegraphics[width=.5\textwidth]{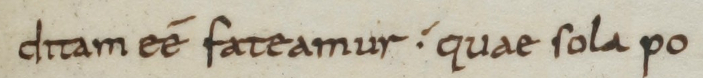}
    \\\hline
No Psych &ditam ee\_ fateamur ; quae soli po cortu\\\hline
Psych &ditam ee\_ fateamur quae sola po
\\\hline
GT&ditam ee\_ fateamur quae sola po
\\\hline

\end{tabular}
\vspace{1mm}
\caption{Sample outputs from the Medieval Latin Manuscripts collected from e-codices that make up our Latin dataset.}
\label{tab:sample_outputs_latin}
\end{table*}